\documentclass[12pt]{article}
\usepackage{a4wide}
\usepackage[scaled]{helvet}
\usepackage[T1]{fontenc}
\usepackage{amsmath}
\usepackage{amssymb}
\usepackage{amsfonts}
\usepackage{tikz}
\usepackage{natbib}
\usepackage{color}
\usepackage{bm}
\usepackage{graphics,graphicx}
\usepackage{url}
\usepackage{hyperref}
\usepackage{pdfpages}
\usepackage{multirow}
\usepackage{tablefootnote}
\usepackage{etaremune}
\usepackage{booktabs}
\usepackage[symbol]{footmisc}
\usepackage{tabularx} 
\usepackage{enumitem}
\SetLabelAlign{CenterWithParen}{\hfil(\makebox[1.0em]{#1})\hfil}

\newcommand{\para}{\bigskip\noindent}

\newcommand{\be}{\begin{equation}}
\newcommand{\ee}{\end{equation}}
\newcommand{\mcS}{\mathcal{S}}
\newcommand{\mcH}{\mathcal{H}}
\newcommand{\TITLE}[1]{{\center{\bf#1}\vskip30pt}}

\newcommand{\AFFILIATION}[1]{{\small\center {\it #1} \\}}
\newcommand{\AUTHORS}[1] {{\small\center{#1}\\}}

\begin{document}
\TITLE{How important are the genes to explain the outcome - the asymmetric Shapley value as an honest importance metric for high-dimensional features}

\AUTHORS{Mark A. van de Wiel$^{1}$\footnote[1]{Corresponding author: mark.vdwiel@amsterdamumc.nl}, Jeroen Goedhart$^1$, Martin Jullum$^2$, Kjersti Aas$^2$}

\AFFILIATION{$^1$Dept of Epidemiology and Data Science, Amsterdam Public Health research institute, Amsterdam University Medical Centers, Amsterdam, The Netherlands\\
$^2$Norwegian Computing Center, Oslo, Norway
}

\begin{abstract}
In clinical prediction settings the importance of a high-dimensional feature like genomics is often assessed by evaluating the change in predictive performance when adding it to a set of traditional clinical variables. This approach is questionable, because it does not account for collinearity nor known directionality of dependencies between variables. We suggest to use asymmetric Shapley values as a more suitable alternative to quantify feature importance in the context of a mixed-dimensional prediction model. We focus on a setting that is particularly relevant in clinical prediction: disease state as a mediating variable for genomic effects, with additional confounders for which the direction of effects may be unknown. We derive efficient algorithms to compute local and global asymmetric Shapley values for this setting.  The former are shown to be very useful for inference, whereas the latter provide interpretation by decomposing any predictive performance metric into contributions of the features. Throughout, we illustrate our framework by a leading example: the prediction of progression-free survival for colorectal cancer patients.
\end{abstract}


\section{Introduction}
In many clinical prediction settings, high-dimensional genomics data are combined with other, low-dimensional variables, such as disease state, age and gender. In such models, the additional predictive strength of the genes is often disappointingly small. This may lead to the premature conclusion that genomics is irrelevant for the disease. However, the approach to quantify importance by leaving out (groups of) variables and assessing the change in predictive strength has two important shortcomings.

\para
First, it does not account well for correlations. If the genomic variables are highly correlated to the other ones, the latter act as a buffer when removing the genomics from the model, leading to little decline in performance. Still, the two types of variables may be equally important for the outcome. In fact, in low-dimensional settings, plain regression deals with such collinearity by simply spreading the effect across the two variables. For general prediction models, the game theoretic Shapley values \cite[]{lundberg2017unified, aas2021explaining} provide a solution in a similar spirit: effects are spread out over correlated variables to explain the model's predictions. Moreover, it allows quantifying importance of groups of variables, which is very useful in genomics settings \cite[]{jullum2021groupshapley}. While the literature on Shapley values for explaining predictive models is huge, fairly little attention is given to applications in genomics. \cite{watson2022interpretable} reviews some of these applications and pleas for the use of Shapley values in genomics. Throughout, we refer to any individual variable (e.g. `age') or group of variables (e.g. `genes') for which we wish to compute these Shapley values as `feature', while reserving the term `variable' for the individual components used in the prediction model.

\para
The Shapley value framework is flexible and can be used to provide either local or global feature importance. Local importance is specific for each individual and feature: it expresses, for example, how much an individual's age contributes to the prediction of the individual's survival. Such a concept is very natural for prediction models that allow for non-linear and interaction effects, as these cause the importance of a feature to be very dependent on its actual value. Hence, local Shapley values allow us to assess which features are important for whom.
Global importance, on the other hand, is useful for comparing features across individuals. Global Shapley value explanations, here referred to as SAGE \cite[Shapley Additive Global importancE;][]{covert2020understanding} utilize a global performance metric like $R^2$ or C-index to quantify feature importance across individuals. Note that the SAGE-$R^2$ is equivalent to `lmg' \cite[]{gromping2007relative}, indicating that such a global metric is also relevant for (complex) regression models.

\para
Formally, both local and global Shapley values rely on the same game-theoretic foundation: they consider the output of the prediction model for many (by default all) subsets of features and contrasts subsets containing a feature with those that do not. The two versions of Shapley values only differ in what is quantified as the model's output: either simply the prediction itself (local) or a predictive performance metric (global). Several estimators for Shapley values have been proposed \cite[]{covert2021explaining}, which mostly differ in how they deal with the features which are not in the subset when computing a prediction. We focus on estimation of Shapley values by (conditional) marginalization, usually referred to as SHAP \cite[SHapley Additive Predictions;][]{lundberg2017unified, aas2021explaining}. We show this to be more suitable in our mixed-dimensional setting than an alternative estimator which is based on model refitting for subsets of features \cite[]{gromping2007relative, williamson2020efficient}. Further discussion about the various estimators of Shapley values is deferred until its formal definition. The literature uses varying terminology for Shapley value based explanations. Table \ref{summary} summarizes the Shapley-related concepts we use here. A more extensive overview is provided by \cite{covert2021explaining}.

\begin{table}
\begin{center}
\begin{tabular}{|l||l|}
\hline
Shapley value$^{*}$ & Local feature importance (explaining individual prediction) \\
SAGE & Global feature importance (explaining model performance)\\\hline
Symmetric & No directional effects \\
Asymmetric & Directional and undirectional effects\\\hline
SHAP & Estimation by marginalizing over features not in the subset \\
Refitting & Estimation by refitting models on subsets of features \\\hline
Marginal approach & SHAP assuming independence for marginalization\\
Conditional approach & SHAP using conditional expectation to account for dependence\\\hline
\multicolumn{2}{l}{\small $^{*}$Also used to denote the general concept} \\
\end{tabular}
\caption{Concepts related to Shapley values.
}\label{summary}
\end{center}
\end{table}

\para
A second shortcoming of the aforementioned leave-one-feature-out approach is that causality or temporal ordering is completely ignored. In a genomics setting it is particularly relevant to use `causally-aware' feature importance metrics \cite[]{watson2022interpretable}, as genomic aberrations are often conceived to be close to the root cause. This means that they may have indirect effects on the outcome via some of the other features. Throughout this manuscript we work with a simple graphical model, depicted in Figure \ref{fd2}, that we believe to be relevant for many diseases. Here, we use disease state $D$, e.g. derived from histological information, as a potential mediator of the effect of $G$ (genes) on outcome $Y$. In addition, confounders $C$ need to be accounted for. $G$ contains the high-dimensional genomics information such as gene expression profiles, plus possibly also a low-dimensional summary such as a consensus clustering based on these profiles, as in our motivational example. Note that $D$ is distinguished from $C$ by the ordering w.r.t. $G$. Our aim is to quantify the importance of $G$, $D$ and $C$ for outcome $Y$, accounting for the dependencies depicted in the graphical model. We propose \emph{asymmetric} Shapley values \cite[]{frye2020asymmetric} as a suitable solution for quantifying feature importance in our mixed-dimensional setting. To adapt the asymmetric Shapley values to this setting, we develop several methodological and computational extensions to the original proposal.

\para
First, to quantify the importance of $G$, we extend the asymmetric Shapley value to its group version \cite[]{jullum2021groupshapley}, which provides a substantial computational shortcut. Second, as computation of Shapley values is known to be potentially very burdensome, we derive subset-based asymmetric Shapley weights, which enable more efficient computation than the ordering-based ones introduced in \cite[]{frye2020asymmetric}. Third, we develop a solution for applications with many features by presenting a novel approximation based on importance sampling.  Fourth, we develop an efficient approach for modelling conditional feature dependencies in our mixed dimensional setting, which is necessary for estimating conditional Shapley values. Following \cite{frye2020asymmetric}, we use conditional SHAP instead of the more commonly employed marginal SHAP (see Table \ref{summary}), since marginal SHAP ignores dependencies and is therefore inconsistent when features have directional relationships. Our suggested dependency modelling approach reduces the dimension of the high-dimensional genetic component, after which established low-dimensional dependency models for Shapley value computation \cite[]{olsen2024comparative} can be used.
Finally, to complete our framework, we develop two methods to perform inference using local Shapley values. This allows us to test the global importance of any of the features for the outcome of interest in the context of the trained model.

\para
Throughout the manuscript we motivate and illustrate the framework by a leading example: the prediction of progression-free survival for colorectal cancer patients. Our aim is to determine which features are most important for explaining this outcome while accounting for the aforementioned asymmetry between genes and disease state. For this, we illustrate how the decomposition property of SAGE for a performance metric like C-index renders it to be a well-interpretable global feature importance metric. Moreover, we provide inference based on local Shapley values to identify significant features. Finally, we show that asymmetric Shapley values better capture the mediating effect of disease state for genes than their symmetric counterparts do.


\section{Motivating Data Example}

\noindent
Our motivating example is about prognosis of relapse-free survival ($Y$) for colorectal cancer patients using genomics data $G$ alongside confounders $C$ and disease state $D$. The data set \cite[]{Mestres2024} contains data from $N = 845$ patients with 253 events, distributed across four tumor stage categories ($D$). The genomics data $G$ consists of a high-dimensional part: gene expression data $G_1$ and a low-dimensional part: Consensus molecular subtype (CMS; $G_2$). This is an established prognostic molecular clustering feature relating to pathways, mutation rates, and metabolics. As detailed further, we will enrich $G$ by another low-dimensional summary $G_3$ which summarizes $G_1$ in the context of $D$. The confounders $C$ consist of gender ($C_1$), age ($C_2$) and left/right tumor site ($C_3$). More details about the data are given in the Section \ref{results}.

\begin{figure}
\begin{center}
\includegraphics[scale=0.6]{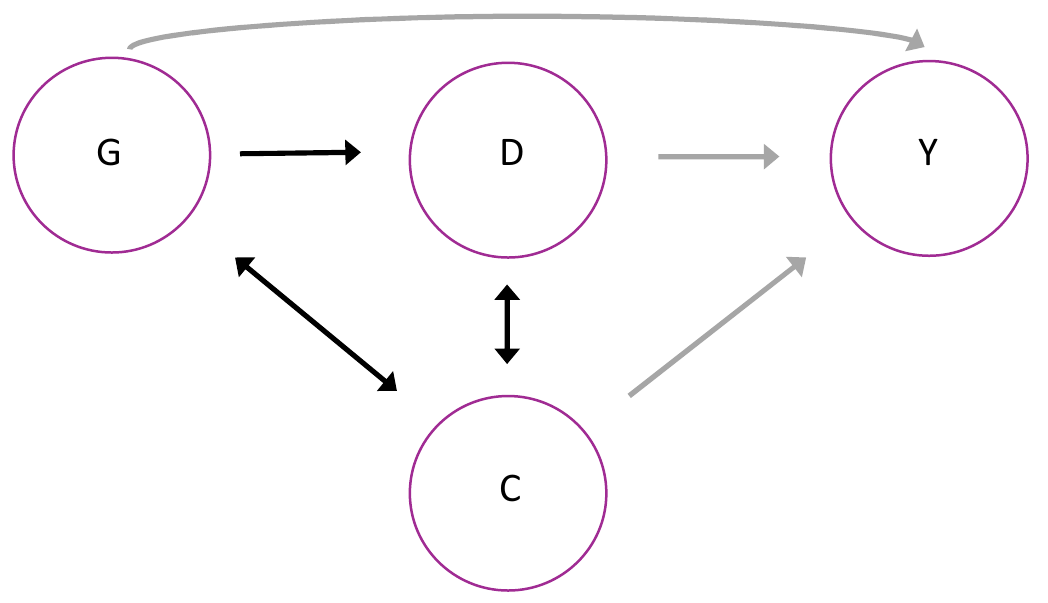}
  \caption{Model for information flow for $G$: gene expression profile, $C$: confounders, $D$: disease state, $Y$: outcome. Right arrows depict unidirectional effects, bidirectional arrows depict correlations. Grey arrows indicate prediction, black arrows dependency between variables.  Modules may contain multiple features.}
 \label{fd2}
 \end{center}
\end{figure}

\para
Our graphical model for ordering $\{G_1, G_2, G_3, C_1, C_2, C_3, D, Y\}$ is depicted in Figure \ref{fd2}, with no additional ordering within modules: $G = \{G_1,G_2, G_3\}$ and $C = \{C_1, C_2, C_3\}$. As this is a mixed-dimensional setting, it is wise to tailor the prediction model such that it handles the low-dimensional variables $\{G_2, G_3, C_1, C_2, C_3, D\}$ differently than the high-dimensional one, $G_1$. Several of such models are discussed by \cite{goedhart2024fusion} and will be considere here. Our goal is to provide interpretation to those models by quantifying the importance of $G$, $C_1, C_2, C_3$ and $D$ for explaining the prediction of $Y$.

\section{Asymmetric group Shapley values}

Our interest lies in quantifying the importance of features that are indexed by $h = 1, \ldots, q$. These features may consist of individual variables, indexed by $j = 1, \ldots, p$, or groups of those.
For clarity we fix the index $i$ for the individual, and let $\bm{x}_h = \bm{x}_{ih} = (x_{ij_1}, \ldots, x_{ij_h})$, with $(j_1, \ldots, j_h)$ denoting the
indices of variables belonging to feature $h$. The realization of $\bm{x}_h$ is denoted by
$\bm{x}^*_h$. Let $\mathcal{H}_{-h}$ be the superset of all subsets of $\{1, \ldots, H\}$ not containing feature $h$. Denote by $\mcS \in \mathcal{H}_{-h}$ the coalition, i.e.~the feature indices for which their value is set at the realization, and its complement $\mcS'$.
Write $\bm{x} = \bigcup_{h=1}^q \bm{x}_h, \bm{x}_{\mcS} = \bigcup_{h \in \mcS} \bm{x}_h, \bm{x}^*_\mcS = \bigcup_{h \in \mcS} \bm{x}^*_h,$ and  $f(\bm{x}) = f(\bm{x}^*_\mcS, \bm{x}_{\mcS'})$ for the predicted value.
Then, the symmetric (group) Shapley value for feature $h$ is defined as:
\begin{equation}\label{phidef}
\phi_h =  \sum_{\mcS \in \mathcal{H}_{-h}} w(\mcS) \biggl(\nu(\mcS \cup \{h\}) - \nu(\mcS)\biggr),
\end{equation}
with $\nu$ being a function measuring the contribution of coalition $\mcS$ in terms of predictions $f(\bm{x}^*_\mcS, \bm{x}_{\mcS'})$. The definition of $\nu$ may be modified to one's need. Several options will be discussed in the next section.

\para
We now modify \eqref{phidef} to allow for asymmetry:
\begin{equation}\label{aphidef}
\phi_h^a =  \sum_{\mcS \in \mathcal{H}^a_{-h}} w_h^a(\mcS) \biggl(\nu(\mcS \cup \{h\}) - \nu(\mcS)\biggr),
\end{equation}
with $\mathcal{H}^a_{-h} \subset \mathcal{H}_{-h}$, containing only coalitions $\mcS$ for which both $\mcS$ and $\mcS \cup \{h\}$ respect the asymmetry, meaning $h_2 \in \mcS \Rightarrow h_1 \in \mcS$ when $h_1 \rightarrow h_2$, and likewise for $\mcS \cup \{h\}.$
Note that \cite{frye2020asymmetric} develop the asymmetric Shapley values differently. They start from the well-known equivalent formulation of symmetric Shapley values, which considers all $H!$ orderings of the features and then adapt this formulation by only allowing orderings that respect the known partial orderings, such as $G \rightarrow D$. Formulation \eqref{aphidef} renders the same results as theirs when using appropriate coalition weights $w_h^a(\mcS)$, as detailed below.  We prefer \eqref{aphidef} over the ordering-based formulation because the former is computationally much more efficient as the number of coalitions (subsets) is vastly outnumbered by the number of orderings. E.g. in the setting with one $G$, one $D$, $r$ confounders $C_1, \ldots, C_r$, and $G \rightarrow D$, the number of allowed orderings equals $(r+2)!/2$, while the number of allowed coalitions equals $2^r*3$. Their ratio equals $26, 7.8*10^4, 1.8*10^9$ for $r=5, 10, 15$, respectively.

\section{Contribution function and estimation}
To be able to compute $\phi_h$ and $\phi_h^a$ one needs to define the contribution function $\nu$, for which the following three aspects need to be considered.

\para
First, global and local explanations require different $\nu$'s. For local explanations, one would like to mimic the prediction with only a subset of the features present. For global explanations, one usually considers a performance metric $\psi_f$ of the prediction model $f$, such as explained variation ($R^2$) or C-index. This global version,  referred to as SAGE \cite[]{covert2020understanding}, then measures the improvement of the prediction when $h$ is added to any of the coalitions, across all (test) individuals.  Both types of explanations will be explored here: global explanation allows an overall comparison between models and features, whereas a local explanation renders a more detailed picture. Moreover, we will connect the two by showing that local Shapley values accommodate post-hoc global inference.


\para
Second, the choice of method for estimating Shapley values impacts $\nu$. The many approaches are discussed in \cite{covert2021explaining}. Here, we briefly discuss two popular classes of varieties: our method of choice, marginalization using SHAP \cite[]{lundberg2017unified, aas2021explaining}, and an alternative based on model refitting. The former defines $\nu$ to be a conditional expectation:
\begin{equation}\label{nu}
\nu(\mcS) = E[f(\bm{x}^*_\mcS, \bm{x}_{\mcS'})|\bm{x}_\mcS = \bm{x}^*_\mcS],
\end{equation}
i.e. the expected value of the prediction function, with the expectation computed over all features $h \in \mcS'$. This definition extends trivially to its global counterpart by substituting $f$ by a performance metric $\psi_f$. Shapley values based on refitting \cite[]{gromping2007relative, williamson2020efficient} use $\nu(\mcS) = f_{\mcS}(\bm{x}^*_\mcS),$ with $f_{\mcS}$ being the prediction model fitted on the coalition $\mcS$.
Both approaches have their pros and cons, and may, in some settings, render similar results \cite[]{lundberg2017unified, reyero2025sobol}, while proven to be equivalent if all models used are optimal \cite[]{covert2021explaining}. The refitting option may seem easier to implement as it only relies on fitting prediction models for the coalitions, but computing time may become excessive when the re-fitting itself is time-consuming. Further, we show that in settings for which (groups of) variables are encoded with different complexities in the prediction model $f$ (e.g. linear for high-dimensional variables and flexible for low-dimensional ones), refitting without changing the formulation of the submodels may render incorrect Shapley values. On the contrary, SHAP handles such mixed-complexity models well. Since such models are very useful for mixed-dimensional settings as in our motivational example \cite[]{goedhart2024fusion}, we focus on SHAP for computing asymmetric Shapley values.

\para
Third, the computation of $\nu$ by \eqref{nu} relies on specifying a dependency model. Note that such dependencies are sometimes ignored, leading to so-called \emph{marginal} Shapley values (as opposed to conditional ones; see discussions in \cite{chen2020true} and \cite{aas2021explaining}). We agree, however, with \cite{frye2020asymmetric} in that asymmetry is not well-compatible with marginal Shapley values: if one accounts for direction, one should also account for correlation. In addition, sparse models are often desirable for genomics applications. For such models, the computationally more efficient marginal Shapley values are not a good approximation of conditional ones \cite[]{chen2020true}.

\section{Effect of asymmetry in the low-dimensional setting}
While the focus in this manuscript is on high-dimensional applications, we briefly discuss the low-dimensional regression setting. This allows for a relatively simple comparison of symmetric and asymmetric Shapley values, both analytically and visually. Supplementary Sections \ref{analow} and \ref{simlow} discuss the results in detail; here we summarize the main findings.

\para
First, we show that in a setting with only two variables, $G$ and $D$, the analytical asymmetric Shapley value for $G$ inherits the effect of $D$ just as it does in standard mediation analysis. When a single confounder $C$ (correlation $\rho$ to both $G$ and $D$) is added, analytical results are more complex, but we still observe a similar effect. Moreover, in such a setting $G$ inherits more of the effect of $C$ than $D$ does, in particular when $\rho$ is large. Second, we visualize the symmetric and asymmetric Shapley values for a simple four-variate ($G, D, C_1, C_2$) simulation with $D$ represented by a quadratic term in the outcome model, while the other terms are linear. Here, we assume $G \rightarrow D$ as in Figure \ref{fd2}. We also compute the global counterpart: SAGE-$R^2$, which quantifies how much of the $R^2$ is explained by each of the variables. Both the local and global Shapley values show a clear increase of the importance of $G$ when we impose asymmetry. For example, $G$ is twice as important as $D$ when quantified by the asymmetric SAGE-$R^2$, while being half as important when its symmetric counterpart is used. Finally, we show that in this setting the refitting version of Shapley values does not correctly transport the quadratic effect of $D$ on $G$, whereas Shapley values based on conditional marginalization (see \eqref{nu}) do.

\section{Computation of asymmetric Shapley weights}
When the number of features, which may represent both single variables and groups of variables, is not very small, we need an efficient method to compute the asymmetric Shapley weights $w_h^a(\mcS)$ in \eqref{aphidef}.
In analogy to the ordinary symmetric Shapley weights, $w_h^a(\mcS)$ is computed by counting the number of allowed orderings that render coalition $\mcS$ and its complement $\mcS'$. Let us first give a simple example.

\para
\emph{Example}. Consider the setting with three features only: $\{G, D, C\}$, with $G \rightarrow D$. Allowed orderings are $GDC, GCD, CGD$, each with ordering weight 1/3 when applying uniform ordering weights; the approach extends trivially to other ordering weights discussed by \cite{frye2020asymmetric}. Then, for computing the asymmetric Shapley value of $D$ ($\phi_{D}^a$) by \eqref{aphidef},  we have only two allowed coalitions that do not contain $D$:
$\mathcal{H}^a_{-D} = \{\{G\}, \{G, C\}\}$. Here, $\emptyset$ and $\{C\}$ are not members of $\mathcal{H}^a_{-D}$, because adding $D$ to either of these two would render an invalid coalition that contains $D$ but not $G$.  The weights of the two allowed coalitions are: $w_D^a(\{G\}) = 1/3$ and $w_D^a(\{G, C\}) = 2/3$, as $D$ is preceded by $G$ alone in one ordering, while it is preceded by both $G$ and $C$ in two orderings. Importantly, unlike in the symmetric setting, the asymmetric weights depend also on $h$, the feature of interest.

\para
For the symmetric setting, conventional Shapley weights for subsets $\mathcal{S}$ are derived from the number of orderings of $q-1$ features (some consisting of multiple variables) that render coalition $\mathcal{S}$. Note that feature $h$ for which the Shapley value is computed is added last to the coalition $\mathcal{S}$, so it does not contribute to the number of orderings. Hence, in a completely symmetric setting the number of orderings equals: $|\mathcal{S}|!(q-1-|\mathcal{S}|)!$.
Combined with the $q!$ weight in the definition of Shapley values based on orderings \cite[]{frye2020asymmetric} this renders the well-known symmetric Shapley weights: $$|\mathcal{S}|!(q-1-|\mathcal{S}|)! / q! =
\frac{1}{q}\binom{q-1}{|\mathcal{S}|}^{-1}.$$
We now follow the same reasoning for asymmetric Shapley weights $w_h^a(\mcS)$. We derive those for ordering structures that are relevant for our type of application. Denote the number of allowed orderings in $\mcS$ by $\#\mcS$, and extend this definition to any set of features. Then, $$w_h^a(\mcS) = \frac{\#\mcS \#(\mcH /(\mathcal{S} \cup \{h\}))}{\#\mcH},$$ as the features in $\mcS$ precede the others, and feature $h$ is added last, so its position is fixed. In line with our graphical model in Figure \ref{fd2}, we assume that the graph for all coalitions $\mcS$ can be decomposed into modules of maximum three classes of features. Then, we compute the asymmetric Shapley weights recursively from the total number of orderings in modules $M = M_j, j=1,2,3; M = \{M_j,M_k\}, (j,k) = (1,2),(1,3),(2,3);$ and $M = \{M_1, M_2, M_3\}$. Here, we may assume without loss of generalization that potential  ordering constraints are always of the type $M_j \rightarrow M_k$ for $j < k$. The following expressions for the number of allowed orderings then suffice for computing $w_h^a(\mcS)$ recursively in our modular setting:

\begin{enumerate}[label=\roman*,align=CenterWithParen]
\item $M=M_j$, $M_j$ symmetric (only correlations): $\#M = \#M_j =|M_j|!$
\item $M=M_j$, $M_j$ completely ordered: $\#M = \#M_j = 1$.
\item $M = \{M_j, M_k\}$, completely symmetric: $\#M = (|M_j| + |M_k|)!$
\item $M = \{M_j, M_k\}$, $M_j \rightarrow M_k$: $\#M = \#M_j \#M_k$
\item $M = \{M_1, M_2, M_3\}$, $M_1 \rightarrow M_2 \rightarrow M_3$: $\#M = \#M_1 \#M_2 \#M_3$
\item $M = \{M_1, M_2, M_3\}$, $M_1 \rightarrow M_2, M_1 \rightarrow M_3$: $\#M = \#M_1 \#M_2 \#M_3 \binom{|M_2| + |M_3|}{|M_2|}$
\item $M = \{M_1, M_2, M_3\}$, $M_1 \rightarrow M_2$:  $\#M_1 \#M_2 \#M_3 \binom{|M_1| + |M_2| + |M_3|}{|M_3|}$ orderings,
\end{enumerate}
with $\#M_j:$ the number of orderings within submodule $M_j$. The binomial coefficient in the latter result is the number of ways to distribute $|M_3|$ `blue balls' among $|M_1| + |M_2| + |M_3|$ balls, once the ordering within each module has been fixed.

\para
\textbf{Example}:\\
Let $\mcH = \{G_1, G_2, D, C_1, C_2, C_3, C_4, C_5, C_6\}$ and impose the ordering constraints $G_1 \rightarrow G_2$, $G = \{G_1, G_2\} \rightarrow D$.  Suppose we aim to compute $w_h^a(\mcS)$ for $h=C_6$ and coalition $\mathcal{S} = \{G_1, G_2, D, C_1, C_2, C_3\}$.
First compute $\#\mathcal{S}$, using expression (vii) above with
$\mathcal{S} = M = \{M_1, M_2, M_3\}$ and
$M_1 = \{G_1, G_2\}, M_2 = \{D\}, M_3 = \{C_1, C_2, C_3\}$.
Then, we have $\#M_1 = 1, \#M_2 = 1, \#M_3 = 3! = 6.$. Moreover, $\binom{|M_1| + |M_2| + |M_3|}{|M_3|} = \binom{6}{3} = 20$. Hence, $\#\mathcal{S} = \#M = 6*20 = 120$. Next we have: $\#(\mcH / (\mathcal{S} \cup \{C_6\})) = \#\{C_4,C_5\} = 2$ and $\#\mcH = 1!1!6!*\binom{9}{6} = 60480$. Therefore, $w_{C_6}^a(\{G_1, G_2, D, C_1, C_2, C_3\}) = 240/60480 =1/252.$

\para
For the computation of $\phi_h^a$ these weights need to be combined with the contributions $\nu(\mcS \cup \{h\}\})$ and $\nu(\mcS)$, see \eqref{aphidef}. For such computations it is convenient to write $\phi_h^a = \phi^{a,+}_h - \phi^{a,-}_h,$ with $\phi^{a,+}_h = \sum w^{a}_h(\mathcal{S}) \nu(\mathcal{S}\cup h)$ and $\phi^{a,-}_h = \sum  w^{a}_h(\mathcal{S})\nu(\mathcal{S})$. Then, we show in Supplementary Section \ref{compashap} that the asymmetric Shapley values for all feature groups and individuals are efficiently computed by adding the results of two matrix multiplications: one for the weights and positive contributions and one for the weights and negative contributions.

\para
When the number of possible coalition subsets becomes too large for complete enumeration, e.g. when the number of confounders in $C$ is large, sampling these according to kernelSHAP weights \cite[]{lundberg2017unified, aas2021explaining} provides an efficient approximation for symmetric Shapley values. Here, the kernel weights are set such that they provide an exact solution when all subsets would be considered. These kernel weights are the same for all features. Unfortunately, due to loss of symmetry, this approach does not work for asymmetric Shapley values. However, the Supplementary Material provides a novel importance sampling scheme that implements efficient sampling of the coalition subsets with probabilities proportional to the the asymmetric Shapley weights $w^a_h(\mathcal{S})$. This enables numerical approximation of $\phi_h^a$ when the number of features is too large to allow enumerating all possible coalitions $\mathcal{S}$.

\section{Modeling dependencies for SHAP-based Shapley values}
For computing SHAP-based (conditional) asymmetric Shapley values one needs to compute $\nu(\mcS) = E[f(\bm{x})|\bm{x}_\mcS = \bm{x}^*_\mcS]$, which requires a model to compute dependencies between variables that are part of $\mcS'$ (hence not of $\mcS$) and the ones in $\mcS$. In our setting, this is particularly complex, because of the mix of high-dimensional variables and low-dimensional ones, including some of categorical (such as gender) or ordinal (disease state $D$) nature. Direct modelling of such dependencies would require careful regularization, which is very complex to model and tune well with such a mixed bag of variables. Instead, we opt to use summary scores of $G_1$, the high-dimensional component of $G$, to make this process much easier and flexible. This allows to make use of all dependency models for estimating $\nu(\mcS)$ in software modules such as \texttt{shapr} \cite[]{jullum2025shapr}.

\para
First, when modelling dependency, we simply replace $G_1$ by $k$ extracted components $\bm{q}(G_1)$, obtained by principal component analysis or by any other feature extraction technique deemed suitable for the data at hand. The usual diagnostics may be performed to choose $k$. If $G_1$ is in coalition $\mcS$, then the dependency modeling of variables in $\mcS'$ is performed conditional on the low-dimenional summary $\bm{q}(G_1)$. If $G_1$ is not in $\mcS$, it needs to be generated for computing the expected prediction $E[f(\bm{x})|\bm{x}_\mcS = \bm{x}^*_\mcS]$. For this, we opt to do so empirically by first sampling the components of $\bm{q} = \bm{q}(G_1)$ conditionally on the variables in $\mcS$, and then replace the sampled $\bm{q}$ by $G_{1,i'}$, where $i' = \text{argmin}_i ||\bm{q}(G_{1,i}) - \bm{q}||_2$.

\para
Second, we compute a `$D$-aware' low-dimensional summary of $G_1$, as the relationship between $G$ and mediator $D$ is particularly important in our setting. Such a summary, denoted by $\bm{q}_D(G_1)$, may be obtained from any learner that predicts $D$ from $G_1$. We recommend to treat $\bm{q}_D(G_1)$ as any of the original features by adding it to the feature set, so that it is used for both prediction of $Y$ and for the dependency model. This may decrease regularization-induced bias \cite[]{hahn2020bayesian} due to the more careful modeling of the effect of $G_1$ via the two directed paths from $G$ to $Y$ (see Figure \ref{fd2}), which may benefit both the predictive performance and the dependency modelling.

\section{Inference}\label{infer}
Conditionally on the trained prediction model, local Shapley values $\phi_{ih}$ of independent test individuals $i = 1, \ldots, n'$ may be regarded as i.i.d. realizations of $\Phi_{h}$: the Shapley value for feature $h$ and a random test individual. As opposed to its global counterpart, local Shapley values provide multiple realizations of its random counterpart, $\Phi_{h}$. This makes them a convenient tool for global inference conditional on the trained model \cite[]{williamson2021, watson2024explaining}. Note that such inference does not propagate the uncertainty of the trained model, which is a harder, model-specific task \cite[]{spadaccini2025hypothesis}.

\para
Marginal inference, which simply correlates $\Phi_{h}$ with test individual outcomes $Y$ without accounting for the other features $k \neq h$, is trivial, as standard statistical testing techniques can be used for this purpose. Here, we propose two alternative tests to evaluate the global importance of feature $h$ (e.g. $G$ or $D$) that do account for the other features. First, when we wish to test feature $h$ we use the decomposition property of Shapley values to write prediction $f_i$ for any test individual $i$ as:
\begin{equation}\label{decom}
f_i = \phi_{ih} + \phi_{i,-h} = \phi_{h} + \sum_{k \neq h} \phi_{ik}.
\end{equation}
This decomposition extends to a random test individual for which $\phi_{i,-h}$ is replaced by its random counterpart, $\Phi_{-h}.$
Then, our global null-hypothesis is:
\begin{equation}\label{null}
H_{0h}: (Y \perp \!\!\! \perp \Phi_{h} | \Phi_{-h}).
\end{equation}


\noindent
We discuss two approaches to test \eqref{null}.
First, when the test sample size is limited, the following semi-parametric approach may be appealing. Regress the test outcome $Y_i$ against $\phi_{ih}$ and $(\phi_{ik})_{k \neq h}$. Then, compute the (partial) likelihood of this model and that of the null-model, which excludes $\phi_{ih}$. This creates likelihoods of two nested models, which may be tested using standard likelihood-ratio testing. Note that this approach is semi-parametric, as it does not assume any parametric relationship between the features and the outcome (the Shapley values encode such potentially very complex associations); it only parameterizes the likelihood for the test outcomes as a linear function of $\phi_{ih}$ and $(\phi_{ik})_{k \neq h}$, which is reasonable given the decomposition property of the prediction \eqref{decom}.
For this approach to be consistent with the learner, the learner should: a) be based on likelihood maximization (or, equivalently, cross-entropy minimization); and b) report Shapley values on the linear predictor scale, as these principles concur with the regression setting used for testing.

\para
Second, and more generically, we may make use of the conditional inference framework \cite[]{Hothorn2006} to test \eqref{null} fully non-parametrically.
Here, we propose a simple approach based on matching. The decomposition of $f_i$ into $\phi_{ih}$ and $\phi_{i, -h}$ is useful for this, as it allows us to straightforwardly apply matching to deal with the conditioning. In short, we propose to match neighboring values of $\phi_{i, -h}$ into pairs or blocks, and use block-wise permutation to test \eqref{null}.
The latter is available in the \texttt{R}-package \texttt{coin} for various types of outcomes. Similar to propensity score matching in causal inference, this approach is based on the rather reasonable assumption that the matching suffices to block the impact of potential dependency between $\Phi_{h}$ and $\Phi_{-h}$ when associating the former with outcome $Y$. It does not require to model the conditional distribution of $\Phi_{h}$ given $\Phi_{-h}$ and applies to any learner or Shapley value, irrespective of the scale of the latter. Of course, this may come at the price of lower power than for the semi-parametric approach above, if the assumptions behind the latter are fulfilled.
The two tests will be illustrated on the data example.

\section{Results}\label{results}

Below we illustrate our framework on the motivational example. We contrast the asymmetric Shapley and SAGE values to their symmetric counterparts and to the conventional `leave-one-feature-out' approach. In addition, we show how SAGE decomposes the C-index to allow for easy interpretation. Moreover, we use Shapley values to infer global importance and to illustrate the mediation effect of the disease state $D$.
\subsection{Data: survival prognosis for colorectal cancer patients}
As a data illustration we use the combined cohorts of colorectal cancer patients available as a single data set in the R-package \texttt{mcsurvdata} (Mestres et al., 2024). Our aim is to explain the prediction of relapse-free survival $Y$. Therefore, patients without survival response are removed from the data set. This renders a final data set of size $N = 845$ and $253$ events. The distribution of number of patients across disease states is: $56, 423, 308, 58$ for tumor stages I, II, III and IV, respectively. Hence, disease state $D$ comprises of a four-level ordinal variable.
Genomic information $G$ consists of a high-dimensional component $G_1$ and a low-dimensional one, $G_2$. Here $G_1$ contains gene expression values of the 500 most variable genes. Of note, results using all 21,292 genes were very similar. As we wish to enable efficient reproducibility of the results, we prefer to use this subset of 500 genes. $G_2$ is a four category nominal variable that contains the consensus molecular subtype (CMS), which is an established molecular clustering based on gene expression. Hence, $G_2$ is a pre-defined low-dimensional summary of the gene expression profile. We consider three confounders: $C_1$: gender; $C_2:$ age;  $C_3$: tumor site (left/right). The models are trained on $N_{\text{train}} = 676$ patients and evaluated on $N_{\text{test}} = 169$ patients.

\para
As discussed before, we augment the set of prediction variables by a disease state specific summary of $G_1$, $\bm{q}_D(G_1)$. For this, we apply a standard random forest (RF) to predict the multiclass outcome $D$ using gene expression variables $G_1$. The out-of-bag predicted disease state probabilities, with the first one removed for redundancy, then define a third, low-dimensional genomic variable: $G_3 = q_D(G_1)$. Here, the use of RF to obtain $G_3$ aligns well with the tree-based dependency model \texttt{ctree}, which we opt to use when computing the contribution function $\nu(\mcS)$ in \eqref{nu} with \texttt{shapr}. All low-dimensional variables, including $G_3$, are used for prediction and dependency modeling. These variables are considered as separate features for evaluating their importance.
All 500 variables in $G_1$ are used for prediction, whereas, as discussed before, we replace $G_1$ by a summary $\bm{q}(G_1)$ to characterize its dependencies with other variables for computing $\nu(\mcS)$. For this, we use 10 principle components extracted from $G_1$, defining $\bm{q}(G_1)$. These explain about 50\% of the variation; increasing the number of components to 20 did not qualitatively change the results. The importance of feature $G_1$ is evaluated as a group. In addition, we also evaluate the importance of all gene-related variables $G = \{G_1, G_2, G_3\}$ together, which is conveniently accommodated by the additivity property of Shapley values.

\subsection{Prediction models}
Before explaining a model, we need to decide what model to focus on. For this, we compare the predictive performances of three models that are designed to handle mixed-complexity data:
\texttt{blockForest} \cite[]{hornung2019block}, \texttt{fusedTree} \cite[]{goedhart2024fusion} and \texttt{ridge0}, fitted by \texttt{glmnet} \cite[]{Friedman2010}. Here, \texttt{blockForest} is a random forest that tunes two sets of hyperparameters: one for the high-dimensional variable, i.e. $G_1$, and one for the rest, representing the low-dimensional block. As such it can adapt to the different dimensions of the two blocks. The learner \texttt{fusedTree} builds a simple decision tree for the low-dimensional variables, thereby allowing for non-linearities and interactions for those variables. Next, the remaining variability in the leaves is modeled by fused ridge regression on $G_1$. Hence, \texttt{fusedTree} prioritizes the low-dimensional variables and models the high-dimensional component with a simpler, regularized parametric model.
Finally,  \texttt{ridge0} is a ridge regression which does not penalize the low-dimensional variables. Cross-validation was used for tuning the hyperparameters.

\para
As performance metrics we use the C-index and time-dependent AUC \cite[]{heagerty2000time} at $t=5$ years, evaluated on the test set.
Table \ref{predperf} shows the results for the three models. We observe that \texttt{blockForest} performs best, with
 \texttt{fusedTree} being a close runner-up and the fully linear model \texttt{ridge0} being inferior. Therefore, we mostly focus on
 \texttt{blockForest} when interpreting results.

\begin{table}[h]
\begin{center}
\begin{tabular}{|l|c|c|}
  \hline
   & C-index & tAUC \\\hline
  \texttt{blockForest}  & 0.754 & 0.789\\
  \texttt{fusedTree}    & 0.747 & 0.777\\
  \texttt{ridge0}       & 0.710 & 0.715\\\hline
   \texttt{blockForest (reduced)}  & 0.731 & 0.737\\
  \hline
\end{tabular}\caption{Predictive performances}\label{predperf}
\end{center}
\end{table}

\subsection{Feature importance}
Let us first employ a traditional approach to evaluate the importance of $G = \{G_1, G_2, G_3\}$ for survival $Y$ by comparing the predictive performance of \texttt{blockForest} with $G$ left out (reduced model) with that of the full model. The reduced model renders C-index: 0.731 and tAUC: 0.737. Hence, leaving out $G$ from the prediction model results in only a very modest drop in performance, compared to 0.754 and 0.789 (see Table \ref{predperf}). Below, we will show that this perceived modest importance of $G$ is misleading, in particular when we account for asymmetry.

\para
We assess the impact of accounting for asymmetry by evaluating feature importance globally on test individuals. For this, we compute the symmetric and asymmetric SAGE-C, obtained from \eqref{phidef} and \eqref{aphidef}, by using $\nu(\mcS) = E[\psi_{\text{c-index}}(\bm{x})|\bm{x}_\mcS = \bm{x}^*_\mcS]$, where $\psi_{\text{c-index}}$ is the C-index. By construction, SAGE-C decomposes the performance of the full \texttt{blockForest} model into contributions of all features. Table \ref{cindshapley} displays the results for both the symmetric and asymmetric SAGE.
We also display the likelihood ratios and the $p$-values from the corresponding test described in Section \ref{infer}, as \texttt{blockForest} satisfies the stated assumptions behind this test. Supplementary Table \ref{cindshapley_cind} shows the $p$-values from the conditional independence test. These show a similar pattern as those from the likelihood ratio test, although the former tend to be larger due to the nonparametric nature of the test.

\begin{figure}
\begin{center}
\includegraphics[scale=0.5]{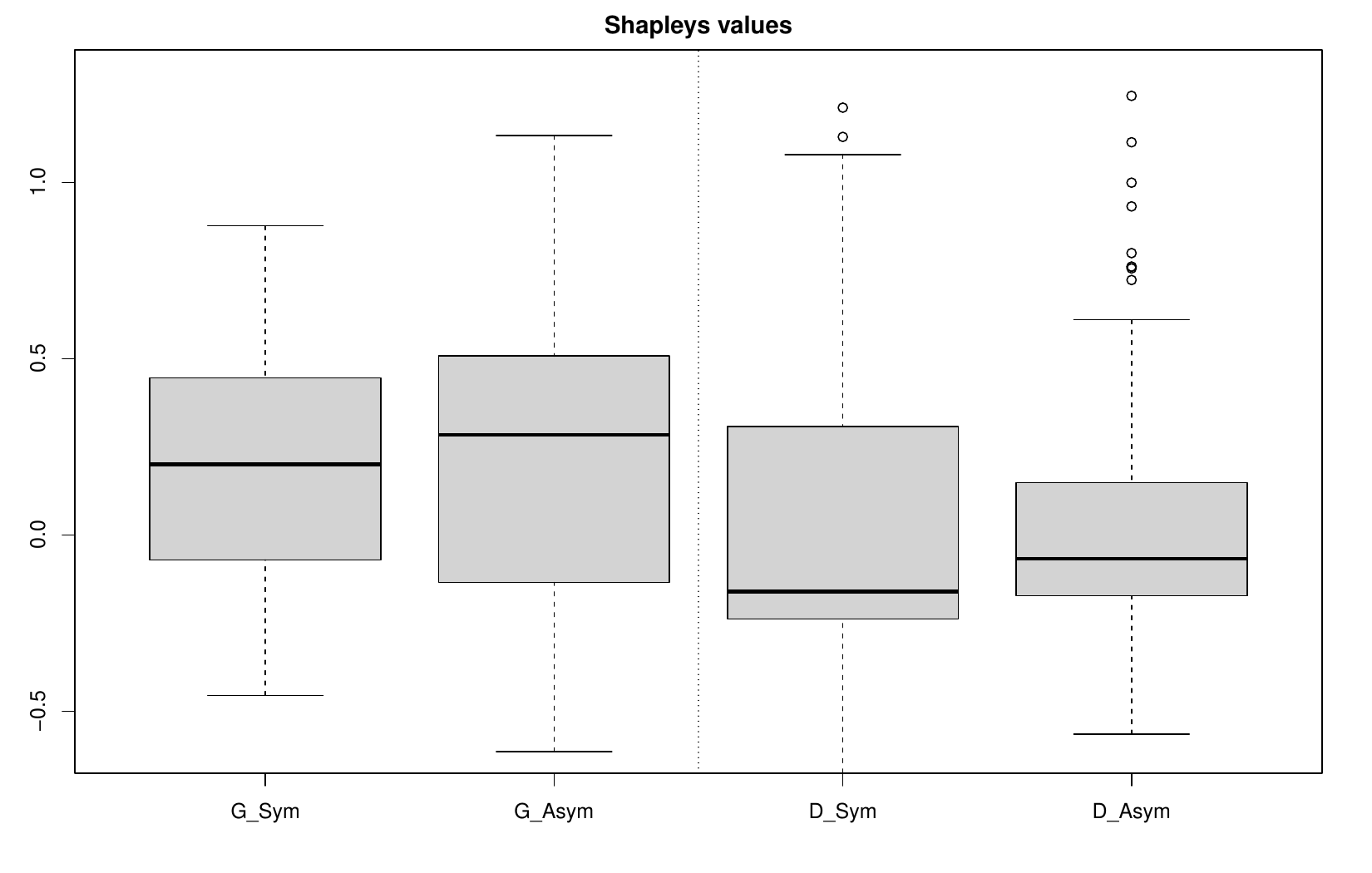}
  \caption{Symmetric and asymmetric Shapley values for $G$ and $D$ for the \texttt{blockForest} model. Shapley values were computed on the log cumulative hazard scale}\label{shapgd}
 \end{center}
\end{figure}

\para
The effect of accounting for asymmetry is clear: $G$ becomes relatively more important at the cost of the importance of $D$. The other two models, \texttt{fusedTree} and \texttt{ridge0}, show similar behavior (Supplementary Table \ref{cindshapley2}).
Moreover, the increased spread in the local Shapley values of $G$ confirm the increased importance of $G$ when accounting for asymmetry (Supplementary Figure \ref{shapgd}). Splitting the contributions of $G$ into that of the high-dimensional component $G_1$ and the low-dimensional one, $G_2 + G_3$, illustrates the importance of adding the low-dimensional components, in particular for the asymmetric version. For this version, the joint Shapley values of $G_2$ and $G_3$ have a significant association with the outcome of the test individuals, which is not the case for the symmetric version. $G_2$ and $G_3$ clearly benefit more from the asymmetry than $G_1$ as they are less strongly regularized than $G_1$ in the \texttt{blockForest} model. In addition, note that the confounders seem to contribute very little to the C-index for both versions. At first, this may be somewhat surprising for `Age', as we study survival. This is likely due to the outcome being progression-free survival instead of overall survival. Despite its small impact on C-index, `Gender' is weakly significant for the symmetric case. This is likely explained by the relative small variability of its Shapley values due to the binary nature of this variable.


\begin{table}[h]
\begin{center}
\begin{tabular}{|l|c|c||r|r|r|r|}
\hline
   &        Sym  &  Asym & $\text{LLR}_{\text{Sym}}$ & $\text{LLR}_{\text{Asym}}$ & $p_{\text{Sym}}$\ \ & $p_{\text{Asym}}$\ \ \\\hline
 intercept & 0.500 & 0.500 &  &  &  &  \\
 Genes ($G$)   &  0.130 &  0.173                                        & 12.39 & 14.56  & 1.72e-05 & 2.12e-06 \\
  --\ \emph{$G$ High} ($G_1$)  & \ \ \ \emph{0.076} & \ \ \ \emph{0.082} & 7.87 & 7.14 & 7.27e-05 & 1.57e-04 \\
  --\ \emph{$G$ Low} ($G_2+G_3$) & \ \ \ \emph{0.053} & \ \ \ \emph{0.091} & 2.28 & 4.91 & 0.102 & 7.39e-03 \\
  --- \ \emph{CMS} ($G_2$)      & \ \ \ \ \ \ \emph{0.028} & \ \ \ \ \ \  \emph{0.043} & 1.59& 2.17 & 0.075 & 0.037 \\
  --- \ \emph{Pred $D$} ($G_3$) & \ \ \ \ \ \ \emph{0.025} & \ \ \ \ \ \ \emph{0.048} & 0.43 & 2.36 & 0.356 & 0.030 \\
  Disease state ($D$)  & 0.091 & 0.062                                  & 13.59 & 10.12 & 1.85e-07 & 6.84e-06 \\
  Gender ($C_1$)       & 0.010 & 0.008                                  & 2.02 & 0.97 & 0.045 & 0.163 \\
  Age ($C_2$)          & 0.021 & 0.010                                  & 0.57 & 0.57 & 0.286 & 0.285 \\
  Tumor site ($C_3$)   & 0.002 & 0.001                                  & 0.05 & 0.02 & 0.760 & 0.843 \\\hline
  Total                & 0.754 &  0.754 & & & &\\\hline \end{tabular}\caption{SAGE decomposition of C-index of \texttt{blockForest} model for symmetric and asymmetric version. Plus (partial) likelihood-ratio (log; LLR) and
$p$-values based on the likelihood-ratio test using Shapley values on log cumulative hazard scale. }\label{cindshapley}
\end{center}
\end{table}

\subsection{Mediation effect}
\begin{figure}
\begin{center}
\includegraphics[scale=0.5]{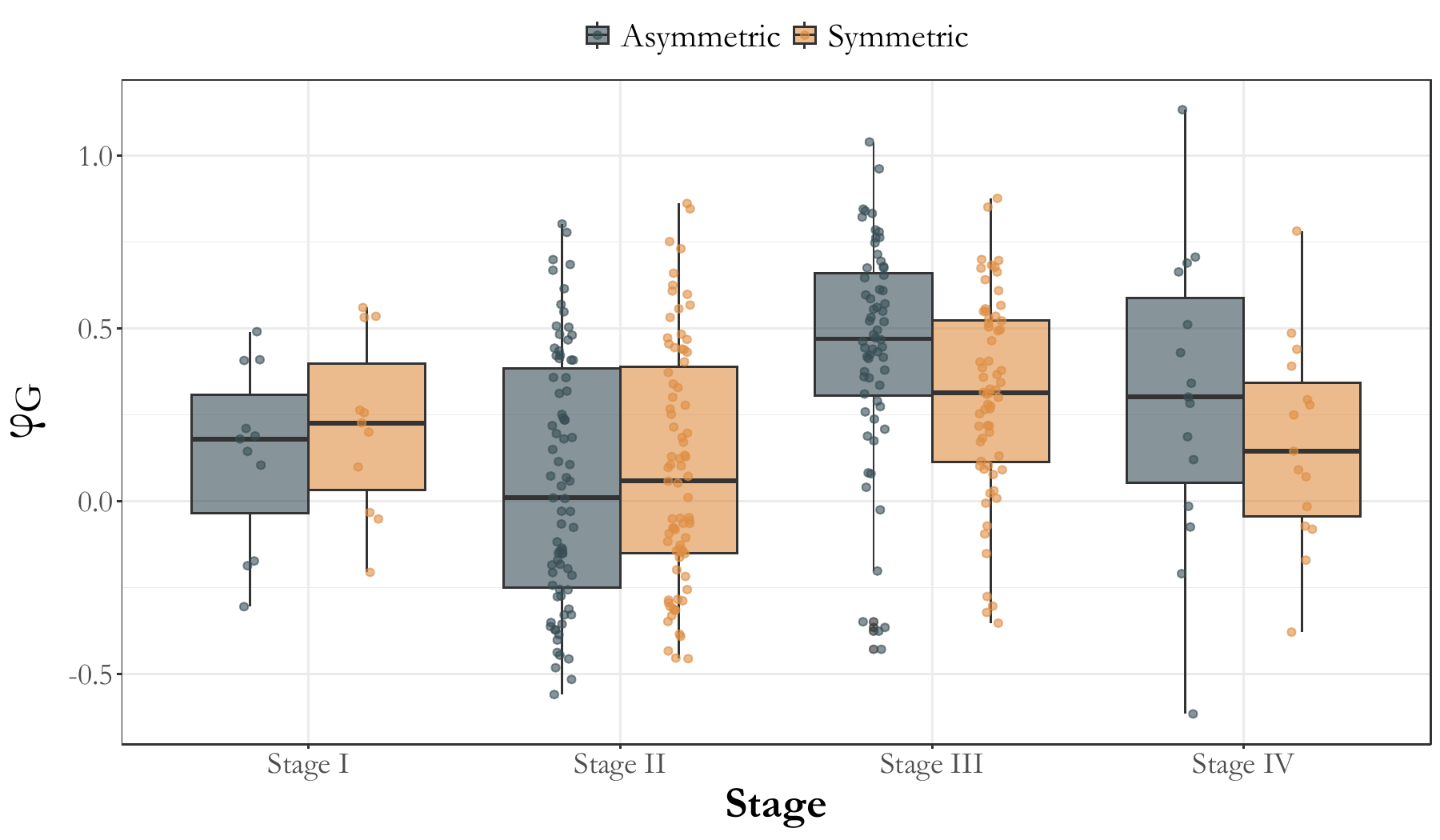}
  \caption{Mediation effect. Shapley values for Genes $G$ (y-axis) against true disease state $D$ (x-axis) for the asymmetric (left box) and symmetric (right box) version.}\label{media}
 \end{center}
\end{figure}

Next, we study the effect of the mediation by disease state $D$ on the local Shapley values of $G$, $\phi_G$. For this, we plot the latter against the true values of $D$ for test individuals. If $D$ would not mediate the effect, then we should not see a difference in the distributions of $\phi_G$. Figure \ref{media} shows a small difference for the symmetric Shapley values and a much larger one for the asymmetric ones. The latter is also much more significant when tested using a Kruskal-Wallis rank test (p = 1.4e-3 vs p = 4.3e-9). The test being significant suggests that tumor stage is associated with changes in how the model relies on high-dimensional omics information in test/unseen patients.

\para
The median difference in $\phi_G$ between the asymmetric and symmetric version becomes more pronounced in stage III and IV patients. This suggests that reallocating part of the predictive contribution of tumor stage to the genes more strongly influences gene importance for patients with advanced disease. All-in-all, Figure \ref{media} shows that the asymmetric Shapley values capture more of the mediation effect of $D$ than their symmetric counterparts.

\section{Implementation}
Our methods have been implemented in \texttt{R}, building on the existing package \texttt{shapr} \cite[]{jullum2025shapr}. We added functions for: 1) efficient computation of the exact asymmetric weights; 2) approximation of these weights by the presented importance sampling algorithm; 3) replacement of the high-dimensional object $G_1$ by a low-dimensional summary $\bm{q}(G_1)$ for dependency modeling (while keeping $G_1$ for predicting $Y$); and 4) inference based on Shapley values. These functions are provided in the \texttt{GeneSHAP} github repository: \url{https://github.com/markvdwiel/GeneSHAP}. This repository also contains the script for reproducing the data example.

\section{Discussion}
We adopted asymmetric Shapley values to quantify feature importance in a mixed-dimensional setting. We showed that a naive leave-one-feature-out approach does not provide a satisfactory solution for this setting. Alternatives that deal better with collinearity \cite[]{verdinelli2024feature} may alleviate the issue, but are not easily amended to deal with asymmetry, as they do not consider multiple subsets.
Causally aware Shapley values \cite[]{heskes2020causal} do account for asymmetry, but have a different aim: quantifying the effect of an intervention. This is a very powerful framework, but it relies on causal knowledge about the bidirectional arcs in Figure \ref{fd2}. These arcs represent marginal correlations with confounders; for computing the effect of an intervention it is essential to know whether these are (merely) cyclic dependencies or caused by an unknown common confounder \cite[]{heskes2020causal}. Such information is often lacking
in clinical prediction settings and reality might be a mixture of the two. Another related framework is mediation analysis \cite[]{schuler2025practical}. It has a somewhat different aim though: estimating the direct and indirect effects of $G$ on $Y$. Again, the direction of the effects between $G$ and confounders $C$ needs to be specified for this. Therefore, these frameworks do not straightforwardly apply to our setting. This also implies a warning about our approach: one should not endow (asymmetric) Shapley values with a causal interpretation without further assumptions.


\para
Our proposed dependency modeling relies on standard dimension reduction techniques to represent the high-dimensional component.
In the context of the data, these techniques can be evaluated by the usual tools, or as part of the dependency modeling in the context of Shapley values \cite[]{olsen2024comparative}. Our motivation for this simple solution for dependency modeling is two-fold. First, for computing conditional  Shapley values, dependency modelling plays a more modest role than in other applications such as synthetic data generation or graphical models. Second, dimension reduction allows for easy use of the available dependency models in \texttt{shapr} \cite[]{jullum2025shapr} without the need for further regularization. Nevertheless, it may be worthwhile to embed specialized dependency models as an alternative into our framework. Candidates might be those used for synthetic data generation, such as \texttt{arf} \cite[]{watson2023adversarial}, once its applicability to mixed-dimensional data has been demonstrated. Alternatively, one could opt to use different dependency models for representing dependencies within the high-dimensional feature, across low-dimensional features and between the high-dimensional feature and the low-dimensional features to account for the differences in complexity.

\para
Another relevant extension is the quantification of feature importance for (many) gene modules like pathways. For this, the challenges are to model dependencies between those modules, and also with the other variables, and to maintain the framework computationally feasible. Again, distinguishing the different feature sets to efficiently model and estimate dependencies may be useful for this purpose.

\para
We discussed two tests for inferring conditional independence of Shapley values from the outcome in a test set: a regression-based test and a non-parametric one. The former may be extended to allow for more interpretation. In particular, one may wish to test whether the Shapley values for the test set are robust \cite[]{joseph2019parametric}, implying the coefficients in the linear decomposition equal one.

\para
Our application focuses on a clinical prediction problem using a high-dimensional genomics feature alongside several confounders and using disease state as a mediator. Our framework may be relevant to some genetics applications as well.
In fact, SAGE-$R^2$ can directly be used as an extension of estimators of the heritability index \cite[]{Visscher, veerman2020estimation}, a popular metric to quantify the importance of genetic variation for explaining phenotypic variation. Traditionally, linear mixed models are used to estimated heritability. Using SAGE-$R^2$ in our framework extends this to non-linear models and to asymmetric settings to account for mediation. Moreover, as shown in the example, our framework allows decomposing the genetics contribution into one of the high-dimensional feature and one of low-dimensional summaries such as the popular polygenic risk score. Including these scores as features in the framework would shed light on the relevance of those in the context of explaining phenotypic variation by a given model.

\newpage
\setcounter{section}{0}
\renewcommand{\thesection}{\Alph{section}}
\section{Supplement: Computing asymmetric Shapley weights}\label{compashap}

\subsection{Shapley weight matrix}
First, note that asymmetric Shapley values \eqref{aphidef} can be written as $\phi_h^a = \phi^{a,+}_h - \phi^{a,-}_h, $ a decomposition that is also used in \cite{kolpaczki2024approximating}.
Here, $\phi^{a,+}_h = \sum_\mathcal{S} w^{a}_h(\mathcal{S}) \nu(\mathcal{S}\cup h)$ and
$\phi^{a,-}_h = \sum_\mathcal{S} w^{a}_h(\mathcal{S})\nu(\mathcal{S})$.
Note here that, by construction, $w^a_h$ constitutes a discrete distribution over all possible $\mathcal{S}$ (hence not containing $h$) for
$\phi^{a,-}_h$. Likewise over all possible $\mathcal{S} \cup h$ for $\phi^{a,+}_h.$ Note that, unlike for symmetric Shapley weights, $w^a_h(\mathcal{S})$ is specific for $\mathcal{S}$ and $h$.

\para
{\bf Example}\\
Define $\mathcal{H}^a$ to be the superset of all coalitions that obey the order restrictions: $\mathcal{H}^a = \bigcup_{h \in H} \mathcal{H}^a_{-h}$. Set $w^{a,-}_h(\mathcal{S}) = w^{a}_h(\mathcal{S})$ if $h \notin \mathcal{S}$ and $w^{a,-}_h(\mathcal{S}) = 0$, otherwise.  Then, for $H = \{G, D, C_1, C_2\}$ and $G \rightarrow D$, we have for $W^{a,-} = \bigl(w^{a}_h(\mathcal{S})\bigr)_{h \in H, \mathcal{S} \in  \mathcal{H}^a}$:

$$W^{a,-} = \frac{1}{12}\left(
    \setlength{\arraycolsep}{0.1cm}
            \begin{array}{c|cccccccccccc}
               & \emptyset & \{ G\} & \{C_1\} & \{C_2\} & \{G, & \{G, & \{G, & \{C_1, & \{G,D, & \{G,D, & \{G,C_1, & \{G,D, \\
                & &  &  &  & D\} & C_1\} & C_2\} & C_2\} & C_1\} & C_2\} & C_2\} & C_1, C_2\}\\\hline
              G & 6 & 0 & 2 & 2 & 0 & 0 & 0 & 2 & 0 & 0 & 0 & 0 \\
              D & 0 & 2 & 0 & 0 & 0 & 2 & 2 & 0 & 0 & 0 & 6 & 0  \\
              C_1 & 3 & 2 & 0 & 1 & 1 & 0 & 2 & 0 & 0 & 3 & 0 & 0 \\
              C_2 & 3 & 2 & 1 & 0 & 1 & 2 & 0 & 0 & 3 & 0 & 0 & 0 \\
            \end{array}
          \right).
$$
\noindent
Similarly, let $w^{a,+}_h(\mathcal{S}) =  w^{a}_h(\mathcal{S}/h)$ if $h \in \mathcal{S}$ and $w^{a,+}_h(\mathcal{S}) = 0$, otherwise. Then, we have for $W^{a,+} = \bigl(w^{a,+}_h(\mathcal{S})\bigr)_{h \in H, \mathcal{S} \in  \mathcal{H}^a}$:

$$W^{a,+} = \frac{1}{12}\left(
 \setlength{\arraycolsep}{0.1cm}
            \begin{array}{c|cccccccccccc}
               & \emptyset & \{ G\} & \{C_1\} & \{C_2\} & \{G, & \{G, & \{G, & \{C_1, & \{G,D, & \{G,D, & \{G,C_1, & \{G,D, \\
                & &  &  &  & D\} & C_1\} & C_2\} & C_2\} & C_1\} & C_2\} & C_2\} & C_1, C_2\}\\\hline
              G & 0 & 6 & 0 & 0 & 0 & 2 & 2 & 0 & 0 & 0 & 2 & 0 \\
              D & 0 & 0 & 0 & 0 & 2 & 0 & 0 & 0 & 2 & 2 & 0 & 6  \\
              C_1 & 0 & 0 & 3 & 0 & 0 & 2 & 0 & 1 & 1 & 0 & 2 & 3 \\
              C_2 & 0 & 0 & 0 & 3 & 0 & 0 & 2 & 1 & 0 & 1 & 2 & 3 \\
            \end{array}
          \right).
$$
Finally, the vector of asymmetric Shapley values $\mathbf{\phi}^a = (\phi_h^a)_{h=1}^q$ is:
$$\mathbf{\phi}^a = (W^{a,+} - W^{a,-})\mathbf{v}^T,$$
with $\mathbf{v} = (\nu(\mathcal{S}))_{\mathcal{S} \in  \mathcal{H}^a}.$

\subsection{Approximating asymmetric Shapley values}
When the number of possible coalition subsets, equalling the number of columns in $W^{a,-}$ and $W^{a,+}$,  is large, exact computation of $\phi_h^a$ is infeasible. Therefore, we need to approximate it.
Due to the asymmetry, it is not trivial to adapt kernelSHAP \cite[]{lundberg2017unified, aas2021explaining} to our setting.
Therefore, we resort to importance sampling as an alternative. First note that $\phi^{-}_h = E_{\bf{w}^{a,-}_h}[\nu(\mathcal{S})],$ with
$\bf{w}^{a,-}_h$: row $h$ of $W^{a,-}$.  Likewise for $\phi^{+}_h$ and $W^{a,+}$. When exact enumeration of all subsets is infeasible,  it is not straightforward to directly sample from $\bf{w}^{a,-}_h$ and $\bf{w}^{a,+}_h$. Therefore, we propose the following importance sampler, which is based on $E_g(X) = E_{g'}[\bigl(g(X)/g'(X)\bigr) X]$ for densities $g$ and $g'$.

\para
{\bf Algorithm}\\
\begin{enumerate}
\item $\tilde{Q}^{-} = \tilde{Q}^{+} = \tilde{\mathbf{w}} = \tilde{v} = \left( \right)$. For $b = 1, \ldots, B:$
\begin{enumerate}
  \item Sample a sequence of feature (groups) that respects the ordering restrictions
  \item Sample a subset by dividing the sequence in two parts with the divisor placed uniformly at one of the $(p+1)$ potential locations. The set before the divisor defines the sampled coalition subset $\mathcal{S}^{(b)}$.
  \item Compute the importance sampling weight: $$w^{(b)}= \frac{1}{p+1}\frac{\#\mathcal{S}^{(b)}\#(\mathcal{H}/\mathcal{S}^{(b)})}{\# \mathcal{H}},$$ with $\#\mathcal{S}:$ the number of ordering-respecting sequences rendering subset $\mathcal{S}$ and $\#\mathcal{H}:$ the total number of
      ordering-respecting sequences.
  \item Compute exact negative and positive Shapley weights for $\mathcal{S}^{(b)}$ for all $p$ features rendering vectors
  $\mathbf{q}^{(b)-}$ and $\mathbf{q}^{(b)+}$ (corresponding to columns of $Q^{-}$ and $Q^{+}).$
  \item $\tilde{Q}^{-} = \left( \tilde{Q}^{-} | \mathbf{q}^{(b)-}\right), \tilde{Q}^{+} = \left(\tilde{Q}^{+} | \mathbf{q}^{(b)+}\right ), \tilde{\mathbf{w}} = \left(\tilde{\mathbf{w}} | w^{(b)}\right )$
  \item $\tilde{\mathbf{v}} = \left(
                 \tilde{\mathbf{v}} | \nu(\mathcal{S}^{(b)}) \right).$
\end{enumerate}
\item The result of this procedure are two $p \times B$ matrices
$\tilde{Q}^{+}$ and $\tilde{Q}^{-}$, and two vectors $\tilde{\mathbf{w}}$ and $\tilde{\mathbf{v}}$ of length $B$. The vector $\hat{\phi}^a$ with $p$ assymetric Shapley values is then given by:
$$\hat{\phi}^a = \bigl(\tilde{W} \odot (\tilde{Q}^{+} - \tilde{Q}^{-})\bigr)\tilde{\mathbf{v}}^T, $$
where $\tilde{W} = \mathbf{1}_p^T \tilde{\mathbf{w}}$ and $\odot$ is the Hadamard product.
\end{enumerate}
\textbf{Notes}\\
\begin{itemize}
\item Sampling sequences in step (a) that respect the ordering restrictions is relatively easy. E.g. when $H = \{G_1, G_2, D, C_1, C_2, C_3, C_4, C_5, C_6\}$ and $G_1, G_2 \rightarrow D$, we first permute all elements of $H$ randomly to render sequence $\pi$. Then, determine the positions of $G_1, G_2, D$ in $\pi$ and replace the first two positions by $\{G_1, G_2\}$ (maintaining their order as it was in $\pi$) and the last by $D$, so that the order restriction is respected.
\item For the importance sampling to be efficient, the weights $w^{(b)}$ should mimic the order of magnitude of the $p$ elements of the true weights, $\mathbf{q}^{(b)-}$ and $\mathbf{q}^{(b)+}$. As for ordinary Shapley values, the latter tend to be large for very small or very large subsets. The uniform sampling of the divisor effectively samples such subsets, which is desirable.
\item Example. For $H = \{G, D, C_1, C_2\}, \mathcal{S}^{(b)} = \{G,D\}$ and $G \rightarrow D$ we have importance sampling weight $w^{(b)} = 1/5* (1*2) / [4! / 2] = 1/30,$ as all sequences in which $D$ preceeds $G$ are not allowed. In this small case, the importance sampling weights for all possible subsets $\mathcal{S}^{(b)}$ can be enumerated:
$$\mathbf{w} = \left(
        \setlength{\arraycolsep}{0.08cm}
            \begin{array}{c|cccccccccccc}
               & \emptyset & \{ G\} & \{C_1\} & \{C_2\} & \{G, & \{G, & \{G, & \{C_1, & \{G,D, & \{G,D, & \{G,C_1, & \{G,D, \\
                & &  &  &  & D\} & C_1\} & C_2\} & C_2\} & C_1\} & C_2\} & C_2\} & C_1, C_2\}\\\hline
              w^{(b)} & 1/5 & 1/10 & 1/20 & 1/20 & 1/30 & 1/15 & 1/15 & 1/30 & 1/20 & 1/20 & 1/10 & 1/5 \\
            \end{array}
          \right).
$$
\item It is straightforward to blend the importance sampling algorithm with exact Shapley weights, computed only for very small and very large subsets. This is computationally feasible as the number of very small and very large subsets is relatively small. Such blending may slightly improve the accuracy of the approximation.
\end{itemize}

\para
The importance sampling algorithm has been implemented and tested independently of the prediction model by setting $\nu(\mathcal{S}) = 1\ \forall \mathcal{S}$. Then, we have for the true values $\phi^{a,-}_h = E_{\bf{w}^{a,-}_h}[\nu(\mathcal{S})] = E_{\bf{w}^{a,-}_h}[1] = 1,$ as $\bf{w}^{a,-}$ and $\bf{w}^{a,-}_h$ constitute probability distributions.  Therefore, we have to verify for all $h$ whether $\hat{\phi}^{a,-}_h = \overline{(\tilde{W} \odot \tilde{Q}^{-})[h,]} \approx 1,$ and likewise for $\hat{\phi}^{a,+}_h$. Table \ref{ISex} shows the accuracy of the approximation using $B=10,000$ sampled subsets for the case with one $G$, one $D$, 12 confounders and $G \rightarrow D$. We observe that all values are indeed close to 1.
\begin{table}[ht]
\centering
 \setlength{\tabcolsep}{0.12cm}
\begin{tabular}{rrrrrrrrrrrrrrr}
  \hline
 & $G$ & $D$ & $C_1$ & $C_2$ & $C_3$ & $C_4$ & $C_5$ & $C_6$ & $C_7$ & $C_8$ & $C_9$ & $C_{10}$ & $C_{11}$ & $C_{12}$ \\
  \hline
$\hat{\phi}^{+}_h$ & 1.01 & 1.00 & 0.96 & 0.98 & 1.01 & 1.02 & 1.01 & 1.01 & 1.00 & 1.01 & 1.00 & 0.99 & 1.03 & 0.98 \\
$\hat{\phi}^{-}_h$ & 0.99 & 1.01 & 1.00 & 1.00 & 1.02 & 1.00 & 1.00 & 1.00 & 1.01 & 0.97 & 1.01 & 1.00 & 0.97 & 0.99 \\
   \hline
\end{tabular}\caption{Estimated asymmetric Shapley values for constant contribution function. Estimation based on importance sampling with $B=10,000$ sampled subsets}\label{ISex}
\end{table}

\newpage
\section{Supplement: Analytical results for low-dimensional toy examples}\label{analow}
To gain intuition on the difference between symmetric and asymmetric Shapley values, we analytically derive exact expressions for two low-dimensional toy examples related to the application of interest. The first example (\autoref{subsec:2D}) only considers a single gene variable ($G$) and a single disease status variable ($\textrm{D}$), for which the expressions are tractable and intuitive. The second example (\autoref{subsec:3D}) also adds a single confounding variable ($C$) and therefore best resembles the application of interest. For the latter example, the expressions become complicated but we provide intuition by deriving limits and showing plots for several scenarios. All derived analytical expressions yield results identical to those obtained from the software implementation.

Recall the Shapley value formula for feature $h$:
\begin{equation}
    \phi_{h}=\sum_{\mathcal{S}\subseteq\{1,\ldots,q\}\setminus\{h\}}w_{\mathcal{S}}\Bigl(\nu\left(\mathcal{S}\cup\{h\}\right)-\nu\left(\mathcal{S}\right)\Bigl),
\label{eq:ShapSetFormula}
\end{equation}
where we suppressed the observation index $i$. Here, $\mathcal{S}$ denotes the subset of features part of the coalition and we denote $\mathcal{S}'$ the complement: the subset of features not part of the coalition. For symmetric Shapley values, all subsets are allowed, while for asymmetric Shapley values subsets that adhere to the causal structure are allowed. For instance, if feature $G$ precedes $D$, then feature $D$ can only be part of the coalition if $G$ is present as well.

The weights $w_{\mathcal{S}}$ compute the number of orderings leading to subset $\mathcal{S}$ divided by the total number of orderings. Computing $w_{\mathcal{S}}$ differs between symmetric and asymmetric Shapley values and is discussed in Section 5 of the main document.

The value function $\nu\left(\mathcal{S}\right)$ is computed by the partial dependence function
$$\nu(\mcS) = E[f(\bm{x}_\mcS, \bm{x}_{\mcS'})|\bm{x}_\mcS = \bm{x}^*_\mcS],$$
rendering the so-called conditional Shapley values \citep{aas2021explaining} (either symmetric or asymmetric) by marginalizing the prediction function over the conditional distribution $p\left(\bm{x}_{\mcS'} \vert \bm{x}_\mcS\right)$. Another commonly employed Shapley value is the marginal Shapley value \citep{lundberg2017unified} which assumes that features are independent: $p\left(\bm{x}_{\mcS'} \vert \bm{x}_\mcS\right)=p\left(\bm{x}_{\mcS'}\right)$. Note that the independence assumption breaks the relationship between the features, rendering identical expressions for the symmetric and asymmetric versions of the marginal Shapley values \citep{frye2020asymmetric}. For comparison, we will show expressions for the (symmetric) marginal Shapley values as well.

\subsection{Two-dimensional toy example}
\label{subsec:2D}
Consider the following set-up
\begin{equation} \label{eq:Toy1}
\begin{split}
G & \sim\mathcal{N}\left(0,1\right)  \\
    D & =\gamma G + \delta, \quad \textrm{where} \quad \delta \sim \mathcal{N}\left({0, \sigma^2_{\delta}}\right) \\
    \hat{f}\left(G,D\right) & = \beta_{1}G + \beta_{2}D.
\end{split}
\end{equation}
Here, $G$ denotes a single gene variable, $D$ a single disease status variable, and we assume we have an estimated model $\hat{f}\left(G,D\right)$. We further assume that $G$ drives $D$ according to the second line of \eqref{eq:Toy1}, rendering the following conditional distribution: $D\vert G \sim \mathcal{N}\left(\gamma G, 1-\gamma^2 \right)$. For simplicity of the expressions, we set $\sigma_{\delta}^2=1-\gamma^2$ to ensure $\textrm{Var}\left(D\right)=1$.

For asymmetric Shapley values, we have to consider the value function contrasts $\nu\left(\{G\}\right)-\nu\left(\{\emptyset\}\right)$ with weight $w=1$ for $G$, and $\nu\left(\{G,D\}\right)-\nu\left(\{G\}\right)$ with $w=1$ for $D$ (as $\nu\left(\{D\}\right)$ is not allowed).

For the symmetric Shapley values, we have to consider the value function contrasts $\nu\left(\{G\}\right)-\nu\left(\{\emptyset\}\right)$ with weight $w=1/2$ and $\nu\left(\{G,D\}\right)-\nu\left(\{D\}\right)$ with $w=1/2$ for $G$ and $\nu\left(\{D\}\right)-\nu\left(\{\emptyset\}\right)$ with $w=1/2$ and $\nu\left(\{G,D\}\right)-\nu\left(\{G\}\right)$ with $w=1/2$ for $D$. Note that the value function computations are identical for the symmetric and asymmetric Shapley values; the only difference is which value function contrasts are allowed.

We now show how to compute $\nu\left(\{G\}\right)$ after which the reader can verify the other value function computations:
\begin{equation} \label{eq:V(1)}
\begin{split}
\nu\left(\{G\}\right) & = \int\hat{f}\left(G,D\right)p\left(D\vert G\right)dD  \\
     & =\int \left(\beta_{1}G + \beta_{2}D\right)\mathcal{N}\left(\gamma G, 1-\gamma^2 \right)dD \\
     & = \beta_{1}G + \beta_{2}\gamma G.
\end{split}
\end{equation}
Similarly to \eqref{eq:V(1)}, we can compute all value functions which results in the Shapley values in Table \ref{tab:Res1}. For comparison, we also show Shapley value expressions assuming independent variables. We leave it to the reader to verify that, for these marginal Shapley values, the choice between the symmetric or asymmetric version yields identical expressions.
\begin{table}[h]
\centering
\caption{Asymmetric, symmetric, and marginal Shapley values ($\phi$) for the toy example described by \eqref{eq:Toy1}.}
\label{tab:Res1}
\begin{tabularx}{\textwidth}{lXXX}
\toprule
\textbf{Variable} & \textbf{Asymmetric} & \textbf{Symmetric} & \textbf{Marginal} \\
\midrule
$G$ & $G\left(\beta_{1}+\beta_{2}\gamma\right)$
    & $\beta_{1}G + \frac{\gamma}{2}\!\left(\beta_{2}G - \beta_{1}D\right)$
    & $\beta_{1}G$ \\

$D$ & $\beta_{2}\!\left(D - \gamma G\right)$
    & $\beta_{2}D + \frac{\gamma}{2}\!\left(\beta_{1}D - \beta_{2}G\right)$
    & $\beta_{2}D$ \\
\bottomrule
\end{tabularx}
\end{table}

From \autoref{tab:Res1}, we can see that the asymmetric Shapley value of $G$ contains the main direct effect ($\beta_1G$) and the indirect effect through $D$ ($\beta_2\gamma G$). Hence, for this simple linear set-up, the asymmetric Shapley value is identical to the total effect obtained from a mediation analysis. Asymmetric Shapley values, however, are more general compared to mediation analysis by being model-agnostic; they allow for complicated nonlinearities and interactions. Furthermore, an arbitrary number of confounders can also be included.

The symmetric Shapley values contain the main effects but also distribute their effect sizes across both $G$ and $D$ due to  the correlation between these variables. Independent Shapley values equal the main effects of the variables as expressed by the model $\hat{f}\left(G,D\right)$.
For all three types of Shapley values, summing the Shapley values across variables recovers the model predictions: $\hat{f}\left(G,D\right) = \beta_{1}G + \beta_{2}D$, as expected.

\subsection{Three-dimensional toy example}
\label{subsec:3D}
Consider the following set-up
\begin{equation} \label{eq:Toy2}
\begin{split}
G & \sim\mathcal{N}\left(0,1\right)  \\
    D & =\gamma G + \delta, \quad \textrm{where} \quad \delta \sim \mathcal{N}\left({0, \sigma^2_{\delta}}\right) \\
    C & =\frac{\rho}{1+\gamma}\left(G+D\right) + \epsilon, \quad \textrm{where} \quad \epsilon \sim \mathcal{N}\left({0, \sigma^2_{\epsilon}}\right) \\
    \hat{f}\left(G,D,C\right) & = \beta_{1}G + \beta_{2}D + \beta_{3}C.
\end{split}
\end{equation}
Here, $G$ again denotes a single gene variable, $D$ a single disease status variable, $C$ a single confounding variable, and $\hat{f}\left(G,D,C\right)$ denotes the estimated model. Set-up \eqref{eq:Toy2} therefore resembles the application of interest.

We again assume that $G$ drives $D$ according to the second line of \eqref{eq:Toy2}. Importantly, we do not assume that $G$ and $D$ drive $C$; the third line of \eqref{eq:Toy2} only ensures that $C$ is correlated both with $G$ and $D$ with correlation strength $\rho$. We set $\sigma_{\delta}^2=1-\gamma^2$ and $\sigma_{\epsilon}^2=1-\frac{2\rho^2}{1+\gamma}$ to ensure that $\textrm{Var}\left(D\right)=\textrm{Var}\left(C\right)=1$, again resulting in simpler expressions for the Shapley values.

The joint density of $(G,D,C)$ equals
\[
(G,D,C) \sim \mathcal{N}\left(\begin{pmatrix}
    0 \\
    0 \\
    0
\end{pmatrix}, \begin{pmatrix}
1 & \gamma & \rho \\
\gamma & 1 & \rho \\
\rho & \rho & 1
\end{pmatrix} \right),
\]
which can be used to compute all conditional distributions. For instance, we have
\[
G \vert D,C \sim \mathcal{N}\left( \frac{\gamma-\rho^2}{1-\rho^2}D + \frac{\rho(1-\gamma)}{1-\rho^2}C,\sigma^2_{G\vert D,C} \right)
\]
and
\[
G,C \vert D \sim \mathcal{N}\left( \begin{pmatrix}
    \gamma D \\
    \rho D
\end{pmatrix},
\begin{pmatrix}
    1-\gamma^2 & \rho(1-\gamma) \\
    \rho(1-\gamma) & 1-\rho^2
\end{pmatrix}
\right).
\]

For the asymmetric Shapley values, we compute the following value function contrasts. For $G$, we compute $\nu\left(\{G\}\right)-\nu\left(\{\emptyset\}\right)$ with weight $w=2/3$ and $\nu\left(\{G,C\}\right)-\nu\left(\{C\}\right)$ with $w=1/3$. For $D$, we compute $\nu\left(\{G,D\}\right)-\nu\left(\{G\}\right)$ with $w=1/3$ and $\nu\left(\{G,D,C\}\right)-\nu\left(\{C,G\}\right)$ with $w=2/3$. For $C$, we compute $\nu\left(\{C\}\right)-\nu\left(\{\emptyset\}\right)$ with $w=1/3$, $\nu\left(\{G,C\}\right)-\nu\left(\{G\}\right)$ with $w=1/3$, and  $\nu\left(\{G,D,C\}\right)-\nu\left(\{G,D\}\right)$ with $w=1/3$. Note that the subsets $\{D\}$ and $\{D,C\}$ are not allowed when calculating asymmetric Shapley values.

For the symmetric Shapley values, we compute the following value function contrasts. For $G$, we compute $\nu\left(\{G\}\right)-\nu\left(\{\emptyset\}\right)$ with weight $w=1/3$, $\nu\left(\{G,C\}\right)-\nu\left(\{C\}\right)$ with $w=1/6$, $\nu\left(\{G,D\}\right)-\nu\left(\{D\}\right)$ with weight $w=1/6$, and $\nu\left(\{G,D,C\}\right)-\nu\left(\{D,C\}\right)$ with weight $w=1/3$. The value function contrasts for $D$ and $C$ are obtained similarly. Note that all subsets are allowed for symmetric Shapley values.

We illustrate how to compute two different value functions for \eqref{eq:Toy2}. We first consider $v\left(\{D\}\right)$:
\begin{equation} \label{eq:V(2,3)}
\begin{split}
\nu\left(\{D\}\right) & = \int\hat{f}\left(G,D,C\right)p\left(G,C\vert D\right)dGdC  \\
     & =\int \left(\beta_{1}G + \beta_{2}D +\beta_{3}C\right)\mathcal{N}\left( \begin{pmatrix}
    \gamma D \\
    \rho D
\end{pmatrix},
\begin{pmatrix}
    1-\gamma^2 & \rho(1-\gamma) \\
    \rho(1-\gamma) & 1-\rho^2
\end{pmatrix}
\right)dGdC \\
     & = \beta_{2}D + \beta_{1}\gamma D + \beta_{3}\rho D.
\end{split}
\end{equation}
and then $v\left(\{D,C\}\right)$:
\begin{equation} \label{eq:V(2)}
\begin{split}
\nu\left(\{D,C\}\right) & = \int\hat{f}\left(G,D,C\right)p\left(G\vert D,C\right)dG  \\
     & =\int \left(\beta_{1}G + \beta_{2}D +\beta_{3}C\right)\mathcal{N}\left( \frac{\gamma-\rho^2}{1-\rho^2}D + \frac{\rho(1-\gamma)}{1-\rho^2}C,\sigma^2_{G\vert D,C} \right)dG \\
     & = \beta_{2}D + \beta_{3}C + \beta_{1}\left[\frac{\gamma-\rho^2}{1-\rho^2}D + \frac{\rho(1-\gamma)}{1-\rho^2}C\right].
\end{split}
\end{equation}

For \eqref{eq:Toy2}, we then have the following asymmetric Shapley values:

\[
\begin{aligned}
\phi^{a}_{G} \;=\;&
G \left[
\beta_1
+ \frac{1}{3} \left( 2 \beta_2 \gamma + 2 \beta_3 \rho
+ \beta_2 \frac{\gamma - \rho^2}{1 - \rho^2} \right)
\right]\\[6pt]
&+ C \left[
\frac{1}{3} \rho \left(\beta_2 \frac{1 - \gamma}{1 - \rho^2} -\beta_1 - \beta_2 \right)
\right]
\\[12pt]
\phi^{a}_{D} \;=\;&
D \left[
\beta_2 + \frac{1}{3} \frac{\beta_3 \rho}{1 + \gamma}
\right]+ \frac{G}{3} \left[-\beta_2 \gamma
- 2\beta_2 \frac{\gamma-\rho^2}{1-\rho^2}
- \frac{\beta_3 \gamma \rho}{1+\gamma}
\right]\\
&- \frac{2C}{3} \left[
\beta_2 \frac{\rho (1 - \gamma)}{1 - \rho^2}
\right]
\\
\phi^{a}_{C} \;=\;&
C\left[
\beta_3
+ \frac{1}{3}\rho(\beta_1+\beta_2)
+ \frac{1}{3}\beta_2\frac{\rho(1-\gamma)}{1-\rho^2}
\right]- \frac{D}{3}\left[\beta_3\frac{\rho}{1+\gamma}\right]\\[6pt]
&+ \frac{G}{3}\left[
\beta_2\frac{\rho^2(\gamma-1)}{1-\rho^2}
- \beta_3\rho
- \beta_3\frac{\rho}{1+\gamma}
\right].
\end{aligned}
\]

The expressions for the exact asymmetric Shapley values are hard to interpret. Summing the asymmetric Shapley values across the variables renders the predictions: $\hat{f}\left(G,D,C\right) = \beta_{1}G + \beta_{2}D + \beta_{3}C,$ as expected. Note that, similarly to the two-dimensional example, the asymmetric Shapley value of $G$ does not depend on $D$. In the limit $\rho=0,$ i.e., the confounder (C) is independent of both the gene (G) and disease state (D) variable, the Shapley value expressions for G and D ($\phi^{a}_{G}$ and $\phi^{a}_{D}$) reduce to the asymmetric ones of the two-dimensional toy example (\autoref{tab:Res1}). In the limit $\gamma=1$, i.e., $G$ completely determines $D$, we have the following expressions:
\begin{equation}
\label{eq:gammaLimit}
\begin{aligned}
\phi^{a}_{G} \;=\;&
G \left[\beta_1+\beta_2 + \frac{2\beta_3 \rho }{3}\right] -C\left[\frac{\rho}{3}(\beta_1 + \beta_2)\right]
\\[12pt]
\phi^{a}_{D} \;=\;&
D\beta_2-G\beta_2=0\\
\\
\phi^{a}_{C} \;=\;&
C\left[
\beta_3
+ \frac{1}{3}\rho(\beta_1+\beta_2)\right]-\frac{G\beta_3 \rho}{3}.
\end{aligned}
\end{equation}
from which one can see that, when $2G\beta_3>C(\beta_1+\beta_2)$, increasing correlation $\rho$ increases the Shapley value for $G$.

For completeness, we also show expressions for the symmetric Shapley values:
\[
\begin{aligned}
\phi^{\textrm{\textbf{sym}}}_{G} \;=\;&
G\left[\beta_1 + \frac{\beta_2 \gamma+\beta_3 \rho}{3} + \frac{\beta_2}{6}\frac{\gamma-\rho^2}{1-\rho^2} +\frac{\beta_3}{6}\frac{\rho}{1+\gamma}\right]\\[6pt]
+&\frac{D}{6}\left[\frac{\beta_3 \rho}{1+\gamma}-\beta_1 \gamma - \beta_3\rho-2\beta_1 \frac{\gamma-\rho^2}{1-\rho^2}\right]\\[6pt]
+&\frac{C}{6}\left[\frac{\beta_2 \rho(1-\gamma)}{1-\rho^2}-\beta_1 \rho - \beta_2 \rho-2\beta_1 \frac{\rho(1-\gamma)}{1-\rho^2}\right]
\end{aligned}
\]

\[
\begin{aligned}
\phi^{\textrm{\textbf{sym}}}_{D} \;=\;&
D\left[\beta_2 + \frac{\beta_1 \gamma+\beta_3 \rho}{3} + \frac{\beta_1}{6}\frac{\gamma-\rho^2}{1-\rho^2} +\frac{\beta_3}{6}\frac{\rho}{1+\gamma}\right]\\[6pt]
+&\frac{G}{6}\left[\frac{\beta_3 \rho}{1+\gamma}-\beta_2 \gamma - \beta_3\rho-2\beta_2 \frac{\gamma-\rho^2}{1-\rho^2}\right]\\[6pt]
+&\frac{C}{6}\left[\frac{\beta_1 \rho(1-\gamma)}{1-\rho^2}-\beta_2 \rho - \beta_1 \rho-2\beta_2 \frac{\rho(1-\gamma)}{1-\rho^2}\right]
\end{aligned}
\]

\[
\begin{aligned}
\phi^{\textrm{\textbf{sym}}}_{C} \;=\;&
C\left[\beta_3 + \frac{\rho(\beta_1+\beta_2)}{3} + \frac{\beta_1+\beta_2}{6}\frac{\rho(1-\gamma)}{1-\rho^2}\right]\\[6pt]
+&\frac{G}{6}\left[\beta_2\frac{\gamma-\rho^2}{1-\rho^2}-\beta_2 \gamma - \beta_3\rho-2\beta_3 \frac{\rho}{1+\gamma}\right]\\[6pt]
+&\frac{D}{6}\left[\beta_1\frac{\gamma-\rho^2}{1-\rho^2}-\beta_1 \gamma - \beta_3 \rho-2\beta_3 \frac{\rho}{1+\gamma}\right].
\end{aligned}
\]
Again, for $\rho=0$, the expressions for $\phi^{\textrm{\textbf{sym}}}_{G}$ and $\phi^{\textrm{\textbf{sym}}}_{D}$ reduce to the symmetric ones of the two-dimensional toy example.
The marginal Shapley values are equal to the main effects as expressed by the model: $\phi^{\textrm{\textbf{marg}}}_{G}=\beta_1G$, $\phi^{\textrm{\textbf{marg}}}_{D}=\beta_2D$, and $\phi^{\textrm{\textbf{marg}}}_{C}=\beta_3C$.

\begin{figure}[h]
\centering{}
\includegraphics[width=\textwidth]{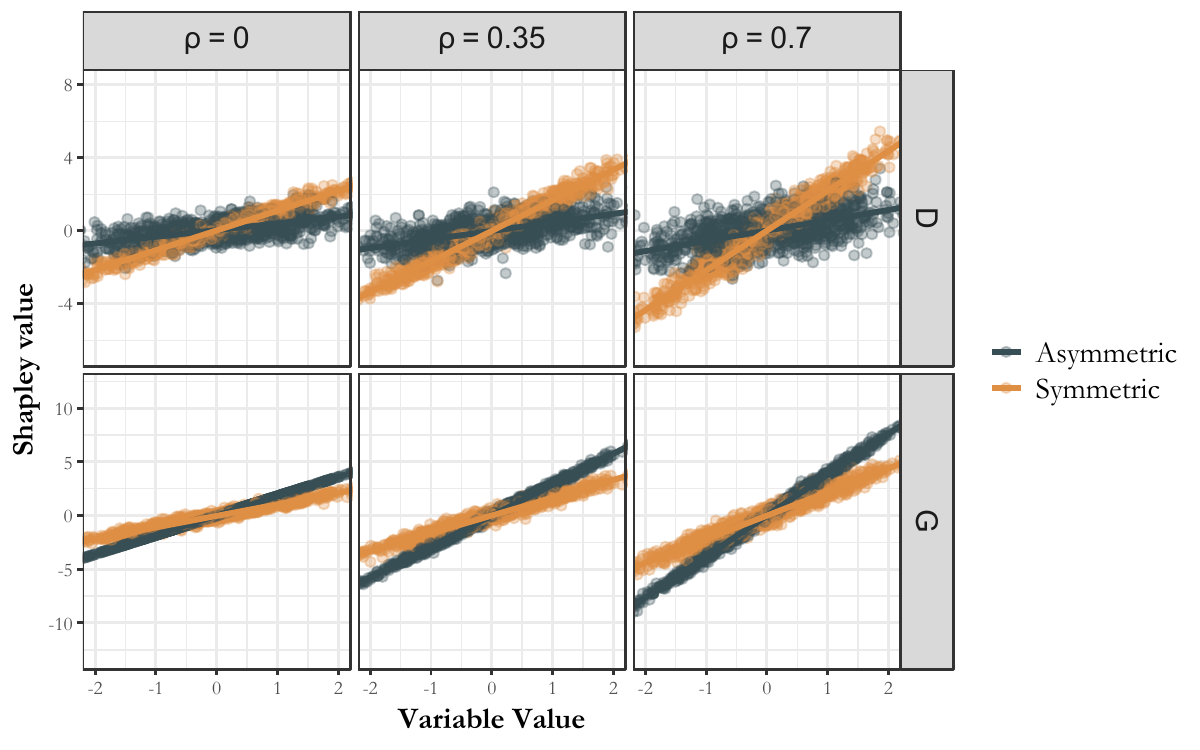}
\caption{Analytical Shapley values for variables $G$ (bottom panels) and $D$ (top panels) as a function of the variable values. Both the symmetric (black) and the asymmetric (orange) versions are depicted. Results are shown for varying correlation strengths $\rho$. Remaining parameter settings: $\beta_1=\beta_2=1.0$, $\beta_3=5.0$, and $\gamma=0.8$.  The plot is produced using the \textsf{R}-package \href{https://cran.r-project.org/web/packages/ggplot2/index.html}{\texttt{ggplot2}} \citep{ggplot2}.}
\label{fig:Sym_vs_Asym}
\end{figure}

\autoref{fig:Sym_vs_Asym} depicts difference between the asymmetric and symmetric Shapley values for \eqref{eq:Toy2}. We plot the expressions for variables $G$ (bottom panels) and $D$ (top panels) for varying strengths of correlation $\rho$ and with fixed parameters $\beta_1=\beta_2=1.0$, $\beta_3=5.0$, and $\gamma=0.8$. Variable values are simulated according to \eqref{eq:Toy2}.

Because $\gamma=0.8$, there is a clear difference in slope between the symmetric and asymmetric Shapley values. Variable $G$ becomes more important in the asymmetric setting whereas $D$ becomes less important. Increasing the correlation strength $\rho$ increases the slopes, in particular for $G$, because $\beta_3$ is set large relative to $\beta_1$ and $\beta_2$ as discussed in \eqref{eq:gammaLimit}. Furthermore, increasing $\rho$ also slightly increases the variability around the slope because the Shapley values will also depend on the comparative value instances of the variables $G$, $D$, and $C$.

\newpage
\section{Supplement: Simulation results for a low-dimensional setting}\label{simlow}
This section illustrates the feature importance metrics in a simple low-dimensional simulation
setting.

\subsection{Simulation set-up}
We show an example in line with the graphical model displayed in Figure \ref{fd2}, with: single gene $G$, continuous disease state $D$, two confounders $C_1$ and $C_2$ and linear outcome $Y$.
\begin{align*}
Y &= \alpha_0 + \alpha_1 G + \alpha_2 C_1 + \alpha_3 C_2 + \alpha_4 D^2 + \epsilon\\
D &= (D_0 + \beta_1 G + U)/\sqrt{6}\\
C_1 &= (C_0 + 2 U)/\sqrt{5}.
\end{align*}
Here, the shared component $U$ models correlation and we assumed $D$ depends only on gene $G$ and correlates only with $C_1$.
We apply the scaling to make sure that $ G, C_1, C_2$ and $D$ are on the same standard Gaussian scale.
We set $\alpha_0 = \alpha_2 = 0,$ $\alpha_1 = \alpha_3 = 1$, $\alpha_4 = \beta_1 = 2$ and generate:
$$
\epsilon, D_0, C_0, U, C_2, G \sim N(0,1),
$$
implying correlations $\rho(D, C_1) = 1/\sqrt{6} * 2/\sqrt{5} = 0.37$.

Therefore, $D$ has a quadratic effect on $Y$, while $G$ and $C_1$ have a linear and $C_2$ has no direct effect on $Y$.
Although one could fit a regression with a quadratic term, we choose to fit a smooth spline to model the effect of $D$ on $Y$ to mimic the situation that we do not know how $D$ relates to $Y$. The model that includes this spline and linear terms for $G, C_1, C_2$ is fitted with \texttt{mgcv}'s  \texttt{gam} function \cite{wood2011fast}, using default parameters. For computing Shapley values, we consider three options. First, marginal Shapley values, assuming independence between the variables. Second, conditional symmetric Shapley values, assuming Gaussian dependency between features, and finally asymmetric Shapley values, assuming $G \rightarrow D$ and Gaussian dependencies.

\subsection{Simulation results}
\begin{figure}
\begin{center}
\includegraphics[scale=0.7]{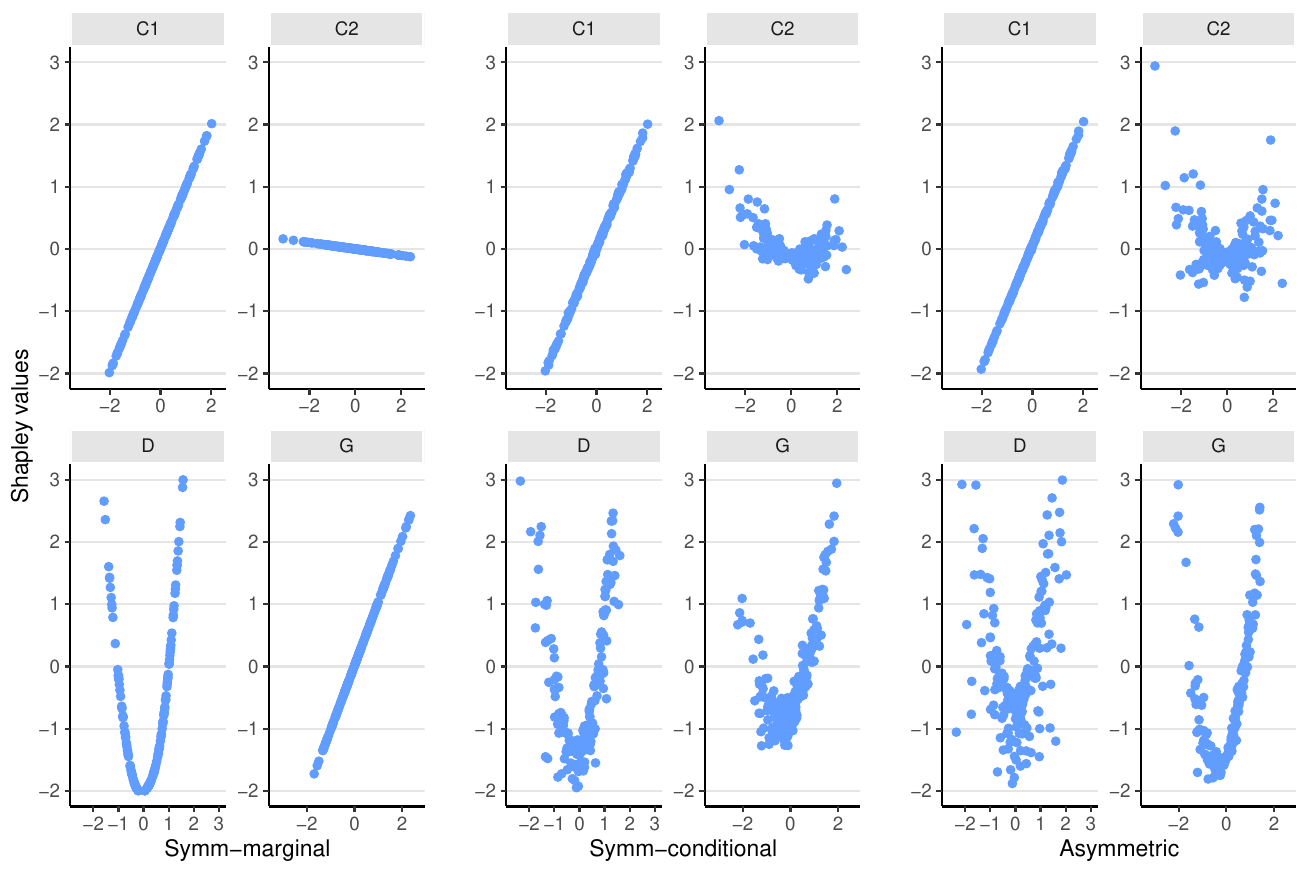}
  \caption{Shapley scatter plots for 200 test observations for the low-dimensional simulation. Variable on the x-axis, its Shapley value on y-axis. All Shapley values are estimated using SHAP, i.e. marginalisation.}
 \label{fig:asymlow}
  \end{center}
\end{figure}

Figure \ref{fig:asymlow} shows the marginal symmetric, conditional symmetric and asymmetric Shapley values resulting from the simulation.

First, observe that the marginal Shapley values reproduce the regression relationships, which is in line with the theory \cite[]{aas2021explaining}, but fail to account for indirect effects on $Y$ that are due to the dependencies of $C_2$ and $G$ with $D$. Conditional Shapley values account for such effects, but the curvature for $G$ is stronger for asymmetric (conditional) Shapley values, as $G$ inherits more importance from $D$ in that setting. The latter is particularly clear for
its global counterpart, SAGE-$R^2$ \cite[]{covert2020understanding}. Table \ref{SageR2} shows that $G$ absorbs a larger fraction of the explained variation when accounting for dependency and asymmetry.

\begin{table}[ht]
\centering
\begin{tabular}{|r|rrr|}
  \hline
 & Sym-marg & Sym-cond & Asym \\\hline
  $C_1$ & 0.11 & 0.11 & 0.11 \\
  $C_2$ & 0.00 & 0.02 & 0.04 \\
  $D$ & 0.66 & 0.50 & 0.26 \\
  $G$ & 0.11 & 0.25 & 0.48 \\\hline
  Total & 0.88 & 0.88 & 0.88\\
   \hline
\end{tabular}
\caption{SAGE-$R^2$ on test observations for the low-dimensional simulation. }\label{SageR2}
\end{table}

\subsection{Comparison with refitting}
\begin{figure}
\begin{center}
\includegraphics[scale=0.5]{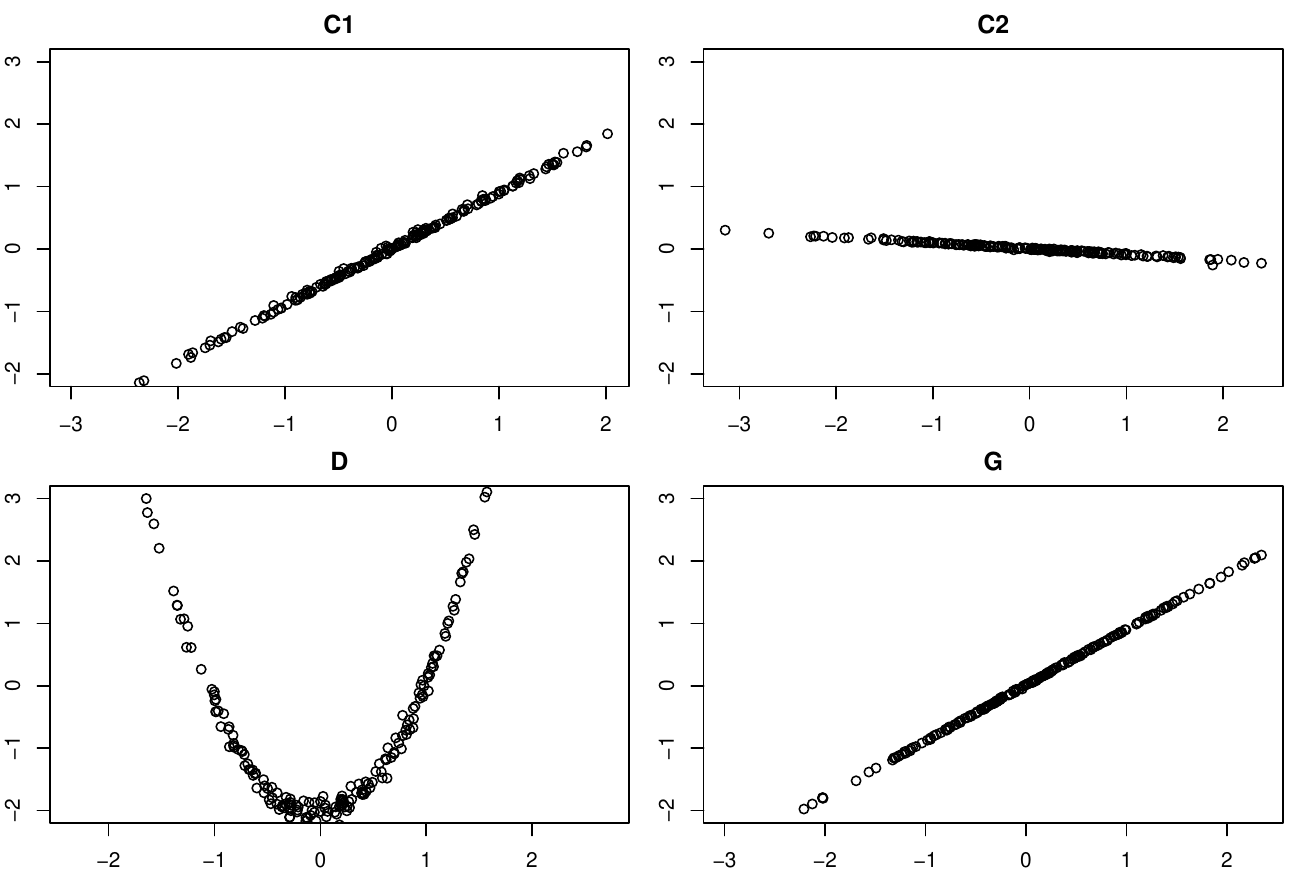}
  \caption{Asymmetric Shapley values based on refitting. Scatter plots for 200 test observations for the low-dimensional simulation. Variable on the x-axis, its Shapley value on y-axis.}
 \label{fig:shap_refit}
 \end{center}
\end{figure}

Finally, Figure \ref{fig:shap_refit} shows the results for asymmetric Shapley values based on refitting submodels for each coalition that obeys $G \rightarrow D$. This figure clearly shows the problem with refitting: the curvature for $G$ (and $C_2$) that should be inherited from $D$ is not picked up by the refitting (cf. right display of Figure \ref{fig:asymlow} for conditional asymmetric Shapley values based on SHAP). This is because the sub-models that do not include $D$ are linear. In fact, the curvature can only be retrieved when we allow a flexible functional form for all other variables in the sub-models. This is generally not desirable, as it will increase variability.

\newpage
\section{Supplement: additional data analysis results}\label{datasup}
\begin{table}[h]
\begin{center}
\begin{tabular}{|l|c|c|c|c|}
\hline
& \multicolumn{2}{|c|}{\texttt{fusedTree}}  & \multicolumn{2}{|c|}{\texttt{ridge0}}\\
   &                      Sym  &  Asym  & Sym  &  Asym \\\hline
Intercept             &  0.500 &  0.500 & 0.500 &  0.500 \\
Genes ($G$)           &  0.122 &  0.170 & 0.075 &  0.143\\
Disease state ($D$)   &  0.121 &  0.071 & 0.118 &  0.058\\
Gender ($C_1$)        &  0.003 &  0.005 & 0.028 &  0.008\\
Age ($C_2$)           &  0.005 &  0.007 & 0.003 &  0.005\\
Tumor site ($C_3$)    &  0.004 & -0.006 & 0.013 & -0.003\\\hline
Total                 &  0.747 &  0.747 & 0.711 &  0.711\\\hline
\end{tabular}\caption{SAGE decomposition of C-index of \texttt{fusedTree} and \texttt{ridge0} models for asymmetric and symmetric setting}\label{cindshapley2}
\end{center}
\end{table}

\begin{table}[h]
\begin{center}
\begin{tabular}{|l|c|c||r|r|}
\hline
   &        Sym  &  Asym  & $p_{\text{Sym}}^{\text{NP}}$\ \ & $p_{\text{Asym}}^{\text{NP}}$\ \ \\\hline
 intercept & 0.500 & 0.500  &  &  \\
 Genes ($G$)   &  0.130 &  0.173                                        & 4.39e-3 & 5.57e-4  \\
  --\ \emph{$G$ High ($G_1$)}  & \ \ \ \emph{0.076} & \ \ \ \emph{0.082} & 0.058 & 0.039   \\
  --\ \emph{$G$ Low} ($G_2+G_3$) & \ \ \ \emph{0.053} & \ \ \ \emph{0.091} & 0.502 & 0.014 \\
  --- \ \emph{CMS ($G_1$)}      & \ \ \ \ \ \ \emph{0.028} & \ \ \ \ \ \ \emph{0.043} & 0.398 & 0.054  \\
  --- \ \emph{Pred $D$ ($G_2$)} & \ \ \ \ \ \ \emph{0.025} & \ \ \ \ \ \  \emph{0.048} & 0.717 & 0.245   \\
  Disease state ($D$)  & 0.091 & 0.062                                  & 7.80e-4 & 3.15e-4  \\
  Gender ($C_1$)       & 0.010 & 0.008                                  & 0.031 & 0.230   \\
  Age ($C_2$)          & 0.021 & 0.010                                  & 0.984 & 0.884   \\
  Tumor site ($C_3$)   & 0.002 & 0.001                                  & 0.981 & 0.788   \\\hline
  Total                & 0.754 &  0.754 & &\\\hline
 \end{tabular}\caption{SAGE decomposition of C-index of \texttt{blockForest} model for symmetric and asymmetric version. Plus
significance based on the nonparametric (NP) conditional independence test using Shapley values on log cumulative hazard scale. }\label{cindshapley_cind}
\end{center}
\end{table}

\bibliographystyle{C://Synchr//Stylefiles//author_short3}
\bibliography{C://Synchr//Bibfiles//bibarrays}      

\end{document}